%% file: acl_latex.tex
\pdfoutput=1

\documentclass[11pt]{article}

\usepackage[final]{acl}

\usepackage{times}
\usepackage{latexsym}

\usepackage[T1]{fontenc}

\usepackage[utf8]{inputenc}

\usepackage{microtype}

\usepackage{inconsolata}

\usepackage{graphicx}

\usepackage{dsfont}

\usepackage{enumitem}
\usepackage{rotating}

\usepackage{booktabs}
\usepackage{multirow}
\usepackage{tcolorbox}
\usepackage{subcaption}
\usepackage{pifont}
\usepackage{arydshln}

\usepackage{amsthm}
\usepackage{amsmath}
\usepackage{amssymb}
\usepackage{xspace}
\newcommand{\narrowtextsc}[1]{\textls[-50]{\textsc{#1}}}
\newcommand{\lm}[1]{\texttt{#1}}
\newcommand{\sys}[1]{\narrowtextsc{#1}}
\newcommand{\data}[1]{\textsf{#1}}

\usepackage{xcolor,colortbl}
\usepackage[dvipsnames]{xcolor}

\usepackage{soul}
\definecolor{lightblue}{RGB}{239,247,250}
\definecolor{blue}{RGB}{224,230,255}
\definecolor{lighpurple}{RGB}{226, 218, 246}
\definecolor{purple}{RGB}{193, 175, 236}
\definecolor{deepred}{RGB}{183,26,26}
\definecolor{deepgreen}{RGB}{4,98,10}
\definecolor{lightred}{RGB}{242,207, 194}

\usepackage[
  separate-uncertainty = true,
  multi-part-units = repeat
]{siunitx}

\usepackage{verbatim}

\usepackage{caption}
\usepackage{subcaption}
\usepackage{booktabs}

\title{Parallel Universes, Parallel Languages: A Comprehensive Study on LLM-based Multilingual Counterfactual Example Generation}

\newcommand{\affilsup}[1]{\rlap{\textsuperscript{\normalfont#1}}}

\author{
    Qianli Wang\affilsup{1,2,\footnotemark[2]}
    \qquad 
    Van Bach Nguyen\affilsup{3}
    \qquad
    Yihong Liu\affilsup{4,5}
    \qquad
    \textbf{Fedor Splitt\affilsup{1}}
    \qquad
    \textbf{Nils Feldhus\affilsup{1,2,6}}
    \\
    \textbf{Christin Seifert}\affilsup{3}
    \qquad
    \textbf{Hinrich Sch\"utze\affilsup{4,5}}
    \qquad
    \textbf{Sebastian M\"oller\affilsup{1,2}}
    \qquad
    \textbf{Vera Schmitt\affilsup{1,2}}
    \\
        $^1$Technische Universit\"at Berlin
    \qquad
        $^2$German Research Center for Artificial Intelligence (DFKI)
    \\
        $^3$University of Marburg 
    \qquad
        $^4$LMU Munich
    \qquad
        $^5$Munich Center for Machine Learning (MCML)
    \\$^6$BIFOLD – Berlin Institute for the Foundations of Learning and Data
        \\
    \normalsize{\footnotemark[2]\textbf{Correspondence}: 
  \texttt{\href{mailto:qianli.wang@tu-berlin.de}{qianli.wang@tu-berlin.de}}
  }
}

\begin{document}
\maketitle
\begin{abstract}
Counterfactuals refer to minimally edited inputs that cause a model's prediction to change, serving as a promising approach to explaining the model's behavior. Large language models (LLMs) excel at generating English counterfactuals and demonstrate multilingual proficiency. However, their effectiveness in generating multilingual counterfactuals remains unclear. 
To this end, we conduct a comprehensive study on multilingual counterfactuals.\footnote{Code and evaluation results are available at: \url{https://github.com/qiaw99/multicfe}} We first conduct automatic evaluations on both directly generated counterfactuals in the target languages and those derived via English translation across six languages. Although translation-based counterfactuals offer higher validity than their directly generated counterparts, they demand substantially more modifications and still fall short of matching the quality of the original English counterfactuals. Second, we find the patterns of edits applied to high-resource European-language counterfactuals to be remarkably similar, suggesting that cross-lingual perturbations follow common strategic principles. Third, we identify and categorize four main types of errors that consistently appear in the generated counterfactuals across languages. Finally, we reveal that multilingual counterfactual data augmentation (CDA) yields larger model performance improvements than cross-lingual CDA, especially for lower-resource languages. Yet, the imperfections of the generated counterfactuals limit gains in model performance and robustness.

\end{abstract}

\section{Introduction}
The importance of providing explanations in multiple languages and illuminating the behavior of multilingual models has been increasingly recognized \cite{cui-etal-2022-multi, zhao-aletras-2024-comparing, resck-etal-2025-explainability, dumas-etal-2025-separating, wang-etal-2025-multilingual-datasets}. 
Counterfactual examples, minimally edited inputs that lead to different model predictions than their original counterparts, shed light on a model's black-box behavior in a contrastive manner \cite{wu-etal-2021-polyjuice, madaan-etal-2022-cfe, zhao-etal-2024-xai}. 
However, despite significant advancements in counterfactual generation methods \cite{ross-etal-2021-explaining, bhan-et-al-2023-tictec, wang-etal-2025-fitcf} and the impressive multilingual capabilities of LLMs \cite{ustun-etal-2024-aya, gao2025thinkingmultilinguallyempowerllm}, these approaches have been applied almost exclusively to English \cite{mcaleese2024comparativeanalysiscounterfactualexplanation, nguyen-etal-2024-llms}. Moreover, cross-lingual analyses have revealed systematic behavioral variations between English and non-English contexts \cite{lai-etal-2023-chatgpt, poelman-lhoneux-2025-roles}, suggesting that English-only counterfactuals are insufficient for capturing the full scope of model behaviors. Nevertheless, the effectiveness of LLMs in generating high-quality multilingual counterfactuals remains an open question. 


\begin{figure*}[!t]
\centering
\resizebox{\textwidth}{!}{
\begin{minipage}{\columnwidth}
\includegraphics[width=\columnwidth]{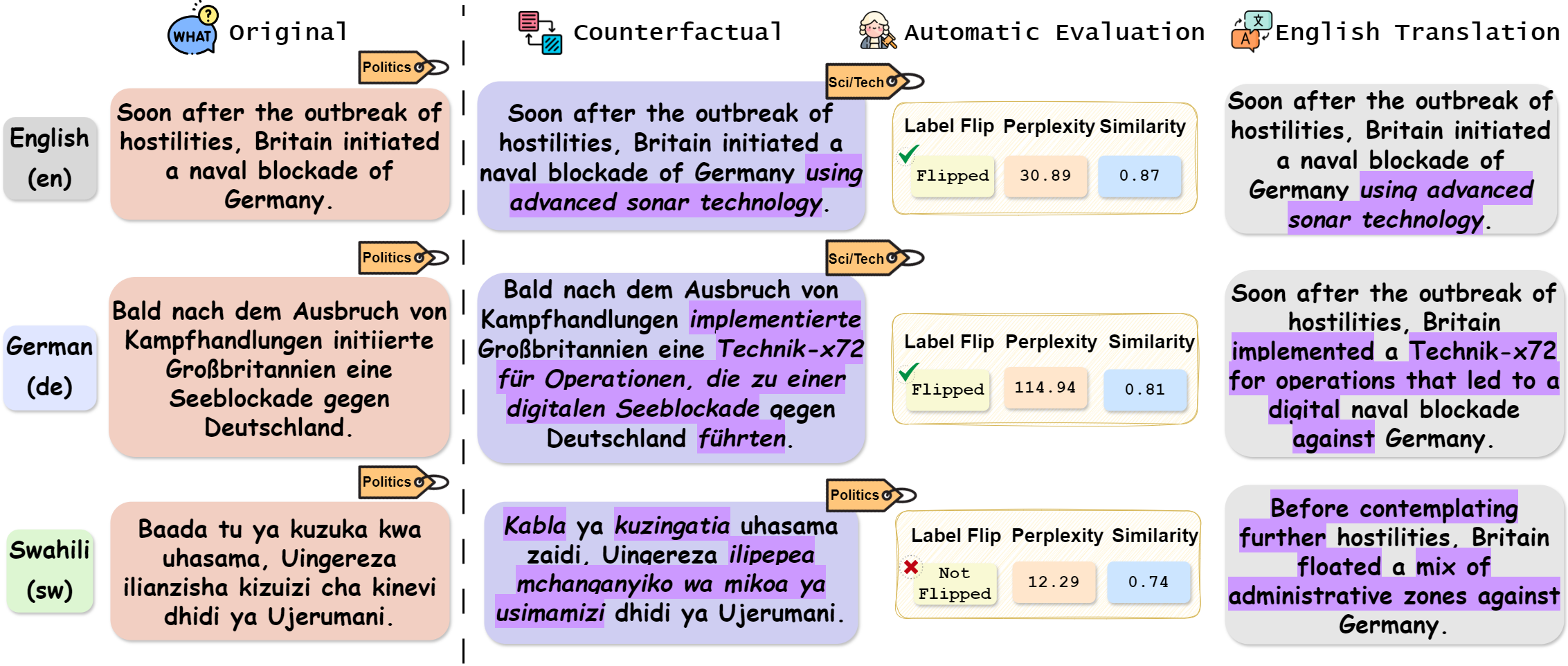}
\end{minipage}
}

\caption{\ding{182} Parallel original inputs from the \data{SIB200} dataset classified as \textbf{``Politics''} in \textit{English} ({\setlength{\fboxsep}{2pt}\colorbox[HTML]{d8d8d8}{\textsf{en}}}), \textit{German} ({\setlength{\fboxsep}{2pt}\colorbox[HTML]{E0E6FF}{\textsf{de}}}), and \textit{Swahili} ({\setlength{\fboxsep}{2pt}\colorbox[HTML]{DDF6D2}{\textsf{sw}}}), \ding{183} their corresponding counterfactuals aimed at changing the label towards ``\textbf{Science/Technology}'' (Sci/Tech), \ding{184} automatic evaluation results and \ding{185} English translations of the generated counterfactuals. Multilingual counterfactuals are evaluated using three automatic metrics (\textit{label flip}$\uparrow$, \textit{perplexity}$\downarrow$ and \textit{similarity}$\uparrow$). In the multilingual counterfactuals and their English translations, words modified by LLMs are highlighted in {\setlength{\fboxsep}{0pt}\colorbox[HTML]{CC99FF}{purple}}.}
\label{fig:example}
\end{figure*}

To bridge this gap, we conduct a comprehensive study on multilingual counterfactuals generated by three LLMs of varying sizes across two multilingual datasets, covering six languages: \textit{English}, \textit{Arabic}, \textit{German}, \textit{Spanish}, \textit{Hindi}, and \textit{Swahili} (Figure~\ref{fig:example}). \textbf{First}, we assess the effectiveness of (1) counterfactuals generated directly in the target language (DG-CFs), and (2) translation-based counterfactuals obtained by translating English counterfactuals (TB-CFs). 
We observe that DG-CFs in high-resource European languages can frequently successfully change the model prediction, as identified by higher label flip rate (LFR). In particular, English counterfactuals generally surpass the LFR of those in other languages. In comparison, TB-CFs outperform DG-CFs in terms of LFR, although they require substantially more modifications. Moreover, TB-CFs show lower LFR compared to the original English counterfactuals from which they are translated. 
\textbf{Second}, we investigate the extent to which analogous modifications are applied in counterfactuals across different languages to alter the semantics of the original input. Our analysis demonstrates that input modifications in English, German, and Spanish exhibit a high degree of similarity; specifically, similar words are edited across languages (cf. Figure~\ref{fig:crosslingual_example}). \textbf{Third}, we report four common error patterns observed in the generated counterfactuals: \textit{copy-paste}, \textit{negation}, \textit{inconsistency} and \textit{language confusion}. Among these, the \textit{copy-paste} issue is the most prominent across different languages. 
\textbf{Lastly}, we investigate the impact of cross-lingual and multilingual counterfactual data augmentation (CDA) on model performance and robustness \cite{liu-etal-2021-counterfactual}. While there are mixed signals regarding performance and robustness gains, multilingual CDA generally achieves better model performance than cross-lingual CDA, particularly for low-resource languages. 


\section{Related Work}
\paragraph{Counterfactual Example Generation.} \sys{MICE} produces contrastive edits that shift a model’s output to a specified alternative prediction \cite{ross-etal-2021-explaining}. Polyjuice leverages a fine-tuned \lm{GPT2} \cite{radford-2019-language} to determine the type of transformation needed for generating counterfactual instances \cite{wu-etal-2021-polyjuice}. \citet{bhan-etal-2023-enhancing} propose a method to determine impactful input tokens with respect to generated counterfactual examples. \sys{CREST} \cite{treviso-etal-2023-crest} generates counterfactual examples by combining rationalization with span-level masked language modeling. \citet{bhattacharjee2024llmguidedcausalexplainabilityblackbox} uncover latent representations in the input and link them back to observable features to craft counterfactuals. \sys{FIZLE} \cite{bhattacharjee-etal-2024-zero} uses LLMs as pseudo-oracles in a zero-shot setting, guided by important words generated by the same LLM, to create counterfactual examples. \sys{ZeroCF} \cite{wang-etal-2025-fitcf} utilizes feature importance methods to pinpoint the important words that steer the generation of counterfactual examples.  However, all of these methods have been evaluated exclusively on English datasets, leaving the ability of LLMs to generate multilingual counterfactuals underexplored.


\paragraph{Counterfactual Explanation Evaluation.}
The quality of counterfactuals can be assessed using various automatic evaluation metrics. \textbf{Label Flip Rate (LFR)} is positioned as the primary evaluation metric for assessing the effectiveness and validity of generated counterfactuals \cite{ross-etal-2021-explaining,ge-2021-counterfactualevaluationexplainableai, nguyen-etal-2024-llms}. LFR is defined as the percentage of instances in which labels are successfully flipped, relative to the total number of generated counterfactuals. \textbf{Similarity} measures the extent of textual modification, typically quantified by edit distance, required to generate the counterfactual \cite{bhattacharjee-etal-2024-zero, wang-etal-2025-fitcf}. \textbf{Diversity} quantifies the average pairwise distance between multiple counterfactual examples for a given input \cite{wu-etal-2021-polyjuice, chen-etal-2023-disco}. \textbf{Fluency} assesses the degree to which a counterfactual resembles human-written text \cite{robeer-etal-2021-generating-realistic, madaan-etal-2022-cfe}. 

\paragraph{Multilingual Counterfactuals.}
\citet{liu-etal-2021-counterfactual} propose using multilingual counterfactuals as additional training data for machine translation -- an approach known as counterfactual data augmentation (CDA). The counterfactuals employed in CDA flip the ground-truth labels rather than the model predictions, and therefore differ from counterfactual explanations explored in this paper. \citet{barriere-cifuentes-2024-study} leverage counterfactuals to evaluate nationality bias across diverse languages. \citet{roy-etal-2025-rag} use counterfactuals to measure answer attribution in a bilingual retrieval augmentation generation system. Nevertheless, none of existing work has investigated how well LLMs are capable of generating high-quality multilingual counterfactual explanations. \looseness=-1


\section{Experimental Setup}
\subsection{Counterfactual Generation}
\label{subsec:cfe_generation}

Our goal is to generate a counterfactual $\tilde{x}$ for the input $x$, changing the original model prediction $y$
to the target label $\tilde{y}$. Our goal is to provide a comprehensive overview of multilingual counterfactual explanations rather than to develop a state-of-the-art generation method. Accordingly, we adopt a well-established counterfactual generation approach proposed by \citet{nguyen-etal-2024-llms}, which is based on \textbf{one-shot Chain-of-Thought} prompting \cite{wei-etal-2022-cot} and satisfies the following properties:
\begin{itemize}
    \item Generated counterfactuals can be used for counterfactual data augmentation (\S\ref{subsec:cda}).
    \item Human intervention or additional training of LLMs is not required, thereby ensuring computational feasibility. 
    \item Generated counterfactuals rely solely on the evaluated LLM to avoid confounding factors, e.g., \textit{extrinsic} important feature signals \cite{bhan-et-al-2023-tictec, wang-etal-2025-fitcf}.
\end{itemize}

LLMs have demonstrated strong multilingual capabilities, yet remain predominantly English-centric due to imbalanced training corpora \cite{jiang2023mistral7b, openai2024gpt4technicalreport, huang2025surveylargelanguagemodels}, resulting in an inherent bias toward English. This imbalance can subsequently hinder the models' proficiency in other languages, often leading to suboptimal performance in non-English contexts \cite{ahuja-etal-2023-mega, zhang-etal-2023-dont, liu-etal-2025-translation}. Consequently, we conduct our experiments using English prompts only.\footnote{Prompt instructions are provided in Appendix~\ref{app:cfe}.}

\begin{figure}[t]
\centering
\resizebox{\columnwidth}{!}{
\begin{minipage}{\columnwidth}
\includegraphics[width=\columnwidth]{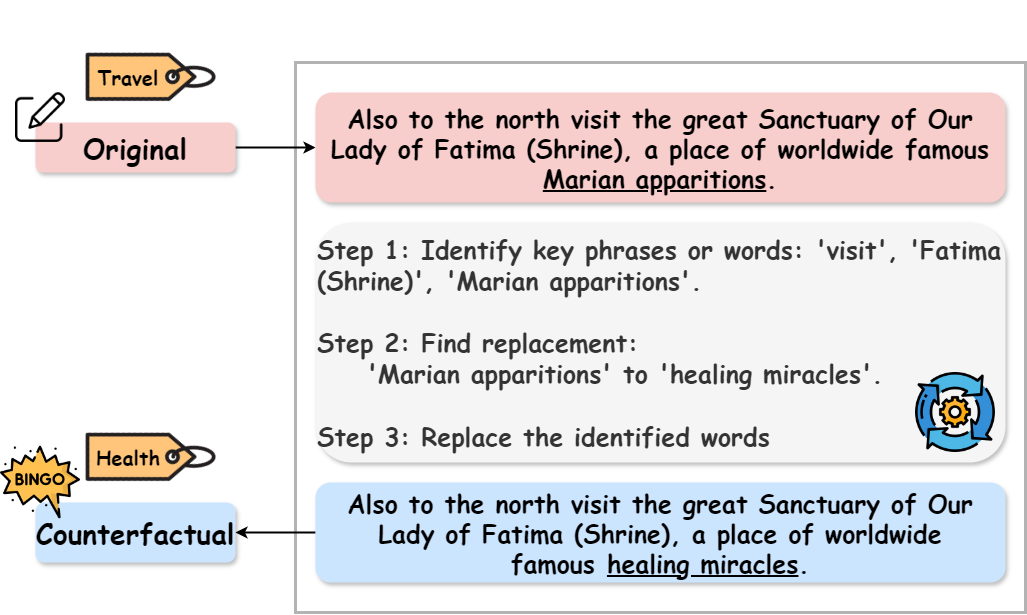}
\end{minipage}
}
\caption{An overview of counterfactual generation process. Given an original instance $x$ from \data{SIB200} classified as ``\textbf{Travel}'', the corresponding counterfactual $\tilde{x}$ is classified as ``\textbf{Health}''. Edits to $x$ are \underline{underlined}.}
\label{fig:instruction}
\end{figure}

We directly generate counterfactuals $\tilde{x}$ (DG-CFs, Table~\ref{subtab:direct}) in target languages through a three-step process as shown in Figure~\ref{fig:instruction}:\looseness=-1
\begin{itemize}
    \item[(1)] Identify the important words in the original input that are most influential in flipping the model's prediction.
    \item[(2)] Find suitable replacements for these identified words that are likely to lead to the target label.
    \item[(3)] Substitute the original words with the selected replacements to construct the counterfactual.
\end{itemize}

Furthermore, we investigate the effectiveness of translation-based counterfactuals $\tilde{x}_{\textsf{en}\text{-}\ell}$ (TB-CFs, Table~\ref{subtab:translation}), where $\ell \in \{\textsf{ar,de,es,hi,sw}\}$. Specifically, LLMs first follow the three-step process in Figure~\ref{fig:instruction} to generate counterfactuals in English. We then apply the same LLM to translate these generated counterfactuals into the target languages (Figure~\ref{fig:translation}). Translation quality is evaluated in Appendix~\ref{app:translation} by automatic evaluation metrics (\S\ref{subsubsec:automatic_evaluation}) and human annotators (\S\ref{subsubsec:user_study}).

\subsection{Datasets}
\label{subsec:datasets}
We focus on two widely studied classification tasks in the counterfactual generation literature: natural language inference and topic classification. Accordingly, we select two task-aligned multilingual datasets and evaluate the resulting multilingual counterfactual examples.\footnote{Dataset examples and label distributions are presented in Appendix~\ref{app:dataset}.}

\paragraph{\data{XNLI}} \cite{conneau-etal-2018-xnli} is designed for cross-lingual natural language inference (NLI) tasks. It extends the English \data{MultiNLI} \cite{williams-etal-2018-broad} corpus by translating into 14 additional languages. \data{XNLI} categorizes the relationship between a \textit{premise} and a \textit{hypothesis} into \textit{entailment}, \textit{contradiction}, or \textit{neutral}.

\paragraph{\data{SIB200}} \cite{adelani-etal-2024-sib} is a large-scale dataset for topic classification across 205 languages. \data{SIB200} categorizes sentences into seven distinct topics: \textit{science/technology}, \textit{travel}, \textit{politics}, \textit{sports}, \textit{health}, \textit{entertainment}, and \textit{geography}. \looseness=-1

\paragraph{Language Selection} We identify six overlapping languages between the \data{XNLI} and \data{SIB200} datasets: \textit{English}, \textit{Arabic}, \textit{German}, \textit{Spanish}, \textit{Hindi}, and \textit{Swahili}. These languages are representative of their typological diversity, spanning a spectrum from widely spoken to low-resource languages and encompassing a variety of scripts.

\subsection{Models}
\label{subsec:models}
We select three state-of-the-art, open-source, instruction fine-tuned LLMs with increasing parameter sizes: \lm{Qwen2.5-7B} \cite{qwen2024qwen25technicalreport}, \lm{Gemma3-27B} \cite{gemmateam2025gemma3technicalreport}, \lm{Llama3.3-70B} \cite{grattafiori2024llama3herdmodels}.\footnote{Further details about the used models for counterfactual generation and inference time can be found in Appendix~\ref{app:experiment}.} These models offer multilingual support and have been trained on data that include multiple selected languages (\S\ref{subsec:datasets}, Appendix~\ref{app:model}). Furthermore, in our experiments, we aim to use counterfactuals to explain a multilingual \lm{BERT} \cite{devlin-etal-2019-bert}, which is fine-tuned on the target dataset (\S\ref{subsec:datasets}).\footnote{Information about the explained models and the training process is detailed in Appendix~\ref{app:explained_model}.}

\section{Evaluation Setup}
\subsection{Automatic Evaluation}
\label{subsec:automatic_evaluation}

We evaluate the generated multilingual counterfactuals using three automated metrics widely adopted in the literature~\cite{ross-etal-2021-explaining, bhan-etal-2023-enhancing, nguyen-etal-2024-ceval-benchmark, wang-etal-2025-fitcf}:

\paragraph{Label Flip Rate (LFR)} quantifies how often counterfactuals lead to changes in their original model predictions \cite{ge-2021-counterfactualevaluationexplainableai, nguyen-etal-2024-llms, bhattacharjee2024llmguidedcausalexplainabilityblackbox}. For a dataset containing $N$ instances, LFR is calculated as:\looseness=-1
\begin{equation*}
    LFR = \frac{1}{N}\sum_{i=1}^{N} \mathds{1} \big(\mathcal{M}(\tilde{x}_i) \neq \mathcal{M}(x_{i})\big)
\end{equation*}
where $\mathds{1}$ is the indicator function, which returns 1 if the condition is true and 0 otherwise. $\mathcal{M}$ denotes the model to be explained. $x_{i}$ represents the original input and $\tilde{x}_i$ is the corresponding counterfactual.

\paragraph{Textual Similarity (TS)} The counterfactual $\tilde{x}$ should closely resemble the original input 
$x$ \cite{madaan-etal-2022-cfe}, with smaller distances signifying higher similarity. Following \citet{bhattacharjee-etal-2024-zero} and \citet{wang-etal-2024-survey}, we employ a pretrained multilingual \lm{SBERT} $\mathbf{e}(\cdot)$\footnote{\url{https://huggingface.co/sentence-transformers/paraphrase-multilingual-MiniLM-L12-v2}} to capture semantic similarity between inputs:\looseness=-1
\begin{equation*}
    TS = \frac{1}{N} \sum_{i=1}^{N} \frac{\mathbf{e}(x_i)^\top \mathbf{e}(\tilde{x}_i)}{\lVert \mathbf{e}(\tilde{x}_{i}) \rVert_2 \, \lVert \mathbf{e}(\tilde{x}_i) \rVert_2}
\end{equation*}

\paragraph{Perplexity (PPL)} is the exponential of the average negative log-likelihood computed over a sequence. It measures the naturalness of text distributions and indicates how fluently a model can predict the subsequent word based on preceding words \cite{fan-etal-2018-hierarchical}. For a given sequence $\mathcal{S} = (t_1, t_2, \cdots, t_n)$, PPL is computed as follows:\looseness=-1
\begin{equation*}
    PPL(\mathcal{S}) = \exp\left\{\frac{1}{n}\sum_{i=1}^{n} \log{p_{\theta}(t_i|t_{<i})}\right\}
\end{equation*}
While \lm{GPT2} parameterized by $\theta$ is commonly used in the counterfactual literature to calculate PPL \cite{le2023cococounterfactuals,nguyen-etal-2024-ceval-benchmark}, it is trained on English data only and is therefore unsuitable for multilingual counterfactual evaluation. Consequently, we use \lm{mGPT-1.3B}\footnote{\url{https://huggingface.co/ai-forever/mGPT}} \cite{shliazhko-etal-2024-mgpt}, which excels at modeling text distributions and provides coverage across all target languages, to compute PPL in our experiments.\footnote{Average perplexity scores of data points in different target languages across each dataset are provided in Table~\ref{tab:perplexity}.}

\subsection{Cross-lingual Edit Similarity}
Following the concept of cross-lingual consistency \cite{qi-etal-2023-cross}, we investigate the extent to which cross-lingual modifications are consistently applied in counterfactuals across different languages to alter the semantics of the original input.\footnote{We instruct LLMs to edit each original input in multiple languages while keeping the target counterfactual label fixed.} To this end, we employ the same multilingual \lm{SBERT} deployed in \S\ref{subsec:automatic_evaluation} to measure the sentence embedding similarity by (1) computing pairwise cosine similarity among directly generated counterfactuals $\tilde{x}_{\ell}$ across different target languages $\ell$; (2) back-translating the directly generated counterfactuals $\tilde{x}_{\ell}$ from language $\ell$ into English $\tilde{x}_{\ell\text{-}\textsf{EN}}$ and quantifying the pairwise cosine similarity among these (back-translated) English counterfactuals.

\subsection{Counterfactual Data Augmentation}
\label{subsec:cda}
To validate whether, and to what extent, counterfactual examples enhance model performance and robustness \cite{Kaushik2020Learning,gardner-etal-2020-evaluating, dixit-etal-2022-core, wang2025truthtwistoptimalmodel}, we conduct cross-lingual and multilingual CDA experiments using a pretrained multilingual \lm{BERT}
. The baseline for CDA is denoted as $\mathcal{M}_{base}$, which is fine-tuned on $D_{\text{base}_{c}}= \left\{\left(x_{i,\textsf{en}}, y_i\right) \mid i \in \mathcal{N}\right\}$ for cross-lingual CDA, and on $D_{\text{base}_{m}}= \left\{\left(x_{i,\ell}, y_i\right) \mid i \in \mathcal{N}, \ell \in \{\textsf{en,ar,de,es,hi,sw}\}\right\}$ for multilingual CDA, where $\mathcal{N}$ denotes the total number of data points. The counterfactually augmented models $\mathcal{M}_{c}$ and $\mathcal{M}_{m}$ are fine-tuned using $D_{\text{base}_{c}}$ and $D_{\text{base}_{m}}$, respectively, along with their corresponding counterfactuals $\tilde{x}_{\ell}$ in the target languages $\ell$, generated either directly (\S\ref{subsec:cda}) or through translation (Appendix~\ref{app:cda}) with different LLMs. 
\looseness=-1

\input{tables/automatic_evaluation}

\section{Results}
\subsection{Multilingual Counterfactual Quality}
\label{subsec:automatic_evaluation}
\subsubsection{Directly Generated Counterfactuals}

Table~\ref{subtab:direct} displays that LFR is dramatically higher for all models on \data{SIB200} than on \data{XNLI}, reflecting the greater inherent difficulty of the NLI task. \textbf{Counterfactuals in English tend to achieve the highest LFR on both \data{XNLI} and \data{SIB200}}. On \data{XNLI}, the gap between high- and low-resource languages widens with model scale, reaching up to $16.46\%$. In contrast, on \data{SIB200}, this gap narrows, where, for instance, counterfactuals in Swahili generated by \lm{Llama3.3-70B} attain the highest LFR. \textbf{Nevertheless, higher-resource European languages (English, German, and Spanish) generally exhibit higher LFRs than lower-resource languages (Arabic, Hindi and Swahili).} Furthermore, counterfactuals in Hindi consistently achieve the best perplexity scores across all three models, indicating superior fluency, whereas counterfactuals in Arabic are generally less fluent. Meanwhile, counterfactuals in Arabic involve more extensive modifications to the original texts indicated by lower textual similarity, whereas those in Swahili and German are generally less edited. However, the higher textual similarity for Swahili reflects fewer LLM edits, resulting in lower LFR. Additionally, no single model produces counterfactuals that are optimal across every metrics and language. \textbf{Likewise, counterfactuals in none of the languages consistently excel across all evaluation metrics.} For example, English counterfactuals achieve higher LFR, but exhibit lower fluency and require more edits than those in other languages, underscoring that the idea of an ``optimal'' or ``suboptimal'' language for counterfactual quality is inherently contextual and metric-dependent.

\subsubsection{Translation-based Counterfactuals}
\label{subsubsec:translation_results}
\paragraph{Comparison with DG-CFs.} Table~\ref{subtab:translation} demonstrates that, in most cases, TB-CFs $\tilde{x}_{\textsf{en}\text{-}\ell}$ yield higher LFR than DG-CFs $\tilde{x}_\ell$ in the target language $\ell$ (Table~\ref{subtab:direct}). In other cases, impairments in translation-based counterfactual quality may suffer from imperfect translations and limitations in LLMs' counterfactual generation capabilities, particularly pronounced on \data{XNLI}. Notably, the LFR improvement is most pronounced for German and least significant for Hindi, although the validity of counterfactuals in Hindi consistently benefits from the translation. Despite TB-CFs $\tilde{x}_{\textsf{en}\text{-}\ell}$ achieving higher LFR compared to DG-CFs $\tilde{x}_\ell$, overall, the LFR of $\tilde{x}_{\textsf{en}\text{-}\ell}$ is lower than that of the original English counterfactuals $\tilde{x}_{\textsf{en}}$. In addition, TB-CFs $\tilde{x}_{\textsf{en}\text{-}\ell}$ are generally less similar to the original input than DF-CFs, showing 15.44\% lower similarity on average. This difference is due to artifacts introduced by machine translation, and they tend to exhibit lower fluency (38\% lower on average) owing to limitations in translation quality. 

\paragraph{Correlation between TB-CFs and Machine Translation.} The degree of LFR improvement is weakly positively correlated with the machine translation quality, measured by automatic evaluation (Spearman's $\rho=0.27$, Table~\ref{tab:automatic_evaluation_translation}) and by human evaluation (Spearman's $\rho=0.07$, Table~\ref{tab:human_evaluation}) (Appendix~\ref{app:translation}). The weak observed correlations suggest that improvements are driven primarily by the quality of the English counterfactuals, with translation quality contributing only to a limited extent.  

    

\begin{figure}[t!]
  \centering

  \begin{subfigure}{\columnwidth}
    \centering
    \centering
\includegraphics[width=\linewidth]{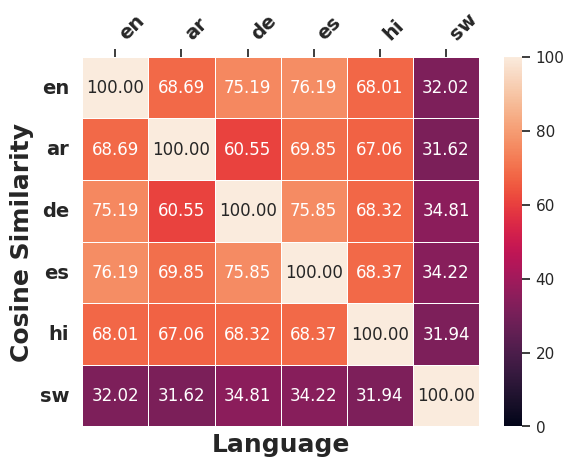}
\caption{\data{XNLI}}
\label{fig:xnli_cos}
  \end{subfigure}
  \hfill
  \begin{subfigure}{\columnwidth}
\centering
\includegraphics[width=\linewidth]{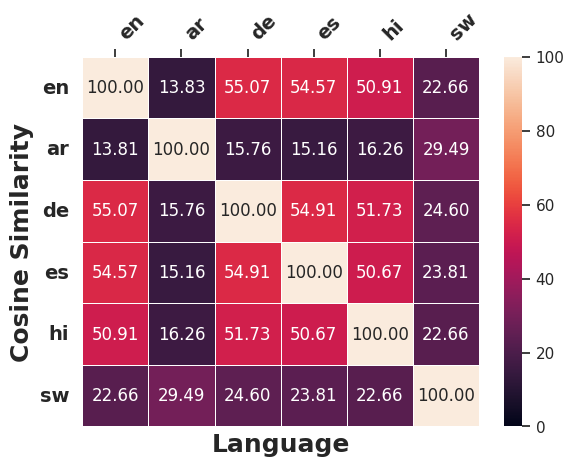}
\caption{\data{SIB200}}
\label{fig:sib_cos}

  \end{subfigure}
  \caption{Cosine similarity scores of counterfactuals $\tilde{x}_\ell$ across different languages measured by \lm{SBERT}.}
  \label{fig:cos_sim_with_en}
\end{figure}

\subsection{Cross-lingual Edit Similarity}
Figure~\ref{fig:cos_sim_with_en} and Figure~\ref{fig:app_cos_sim_trans} indicate that LLMs generally edit inputs for Swahili and Arabic counterfactuals in a substantially different manner than other languages, as evidenced by lower cosine similarity scores.\footnote{Cosine similarity scores for original input and back-translated counterfactuals $\tilde{x}_{\ell\text{-}\textsf{en}}$ in English from \data{XNLI} and \data{SIB200}, are provided in Appendix~\ref{app:crosslingual_sim}.} Notably, for European languages (English, German and Spanish), LLMs tend to apply similar modifications to the original input during counterfactual generation (Figure~\ref{fig:crosslingual_example}), likely because of structural and lexical similarities among these languages \citep{haspelmath2005world,holman2011automated} (Appendix~\ref{app:crosslingual_example}). Additionally, the edits applied across different languages when generating counterfactuals on \data{SIB200} differ markedly from those on \data{XNLI}, as reflected in noticeable differences in cosine similarity scores between the two datasets. This disparity likely stem from \data{SIB200}'s focus on topic classification. When a target label is specified, compared to \data{XNLI}, there might be more distinct ways to construct valid counterfactuals that elicit the required prediction change. \looseness=-1

\begin{figure}[t!]
  \centering

  \begin{subfigure}{\columnwidth}
    \centering
    \centering
\includegraphics[width=\linewidth]{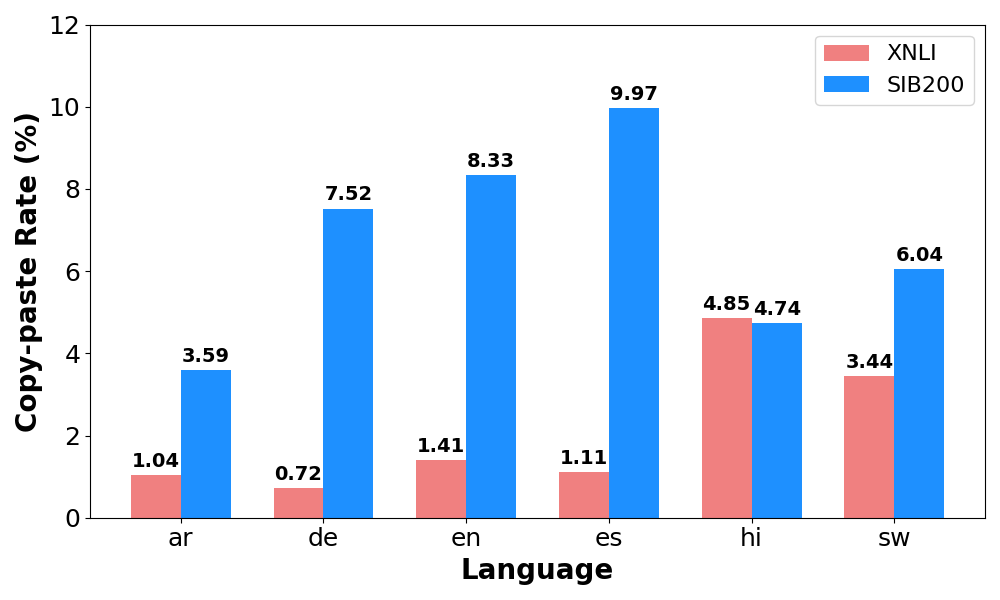}
\caption{Copy-paste rate}
\label{fig:repetitive_rate}
  \end{subfigure}
  \hfill
  \begin{subfigure}{\columnwidth}
\centering
\includegraphics[width=\linewidth]{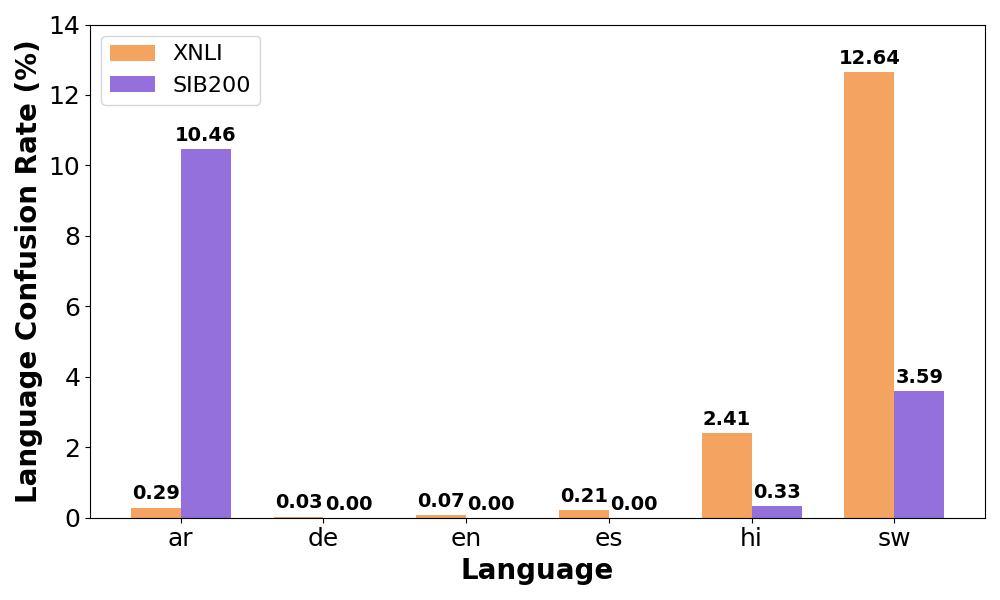}
\caption{Language confusion rate}
\label{fig:lang_conf}

  \end{subfigure}

  \caption{(a) Copy-paste rates and (b) language confusion rates for counterfactuals across different languages.}
  \label{fig:rate}
\end{figure}

\begin{figure*}[t]
\resizebox{\textwidth}{!}{
\begin{minipage}{\textwidth}
\includegraphics[width=\textwidth]{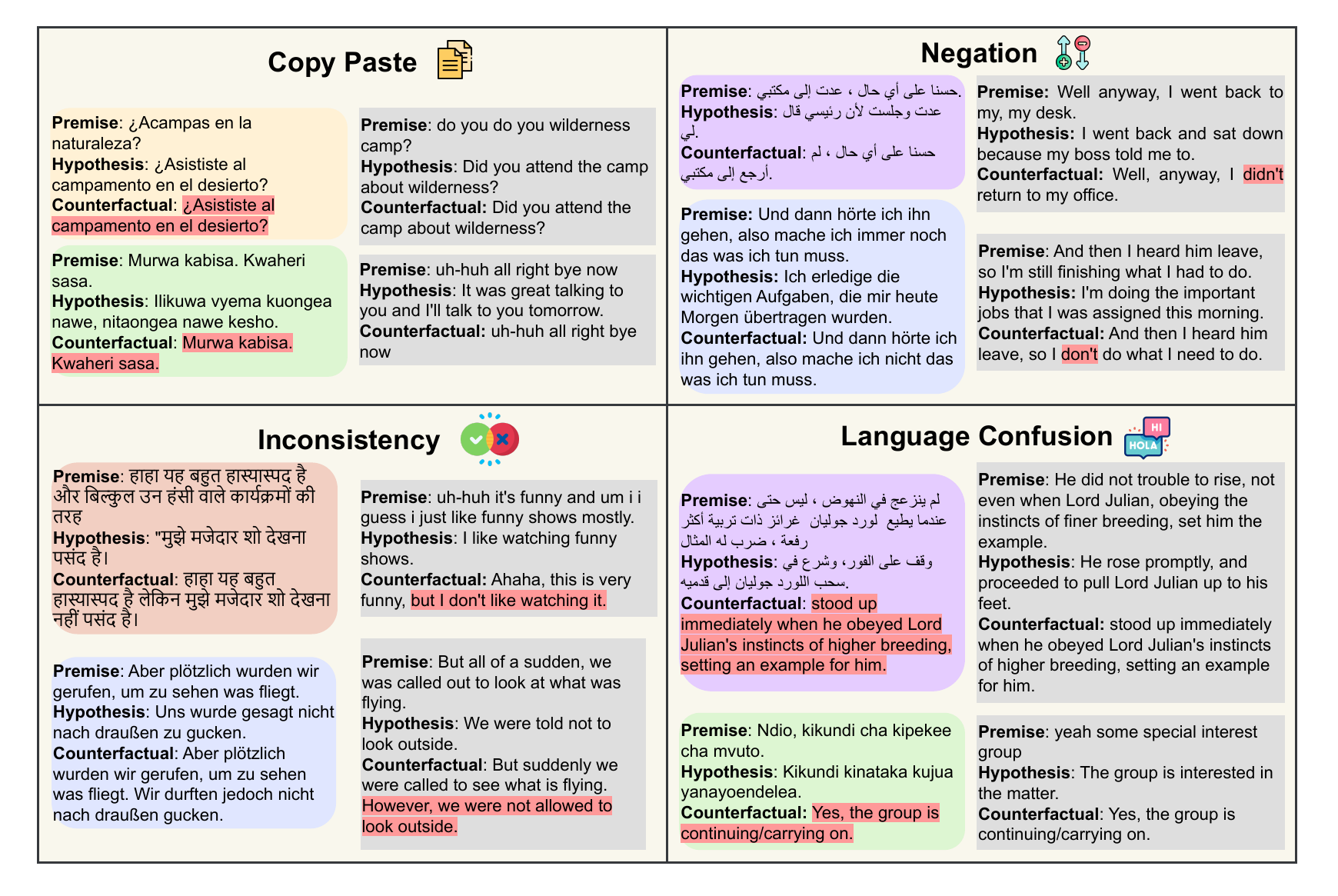}
\end{minipage}
}
\caption{
Categorization of error types in generating multilingual counterfactuals across five languages:
{\setlength{\fboxsep}{1pt}\colorbox[HTML]{E5CCFF}{\textit{Arabic}}}, {\setlength{\fboxsep}{1pt}\colorbox[HTML]{E0E6FF}{\textit{German}}}, {\setlength{\fboxsep}{1pt}\colorbox[HTML]{FFF1D5}{\textit{Spanish}}}, {\setlength{\fboxsep}{1pt}\colorbox[HTML]{F2CFC2}{\textit{Hindi}}}, and {\setlength{\fboxsep}{1pt}\colorbox[HTML]{DDF6D2}{\textit{Swahili}}}.
For each error type, we present two examples and their corresponding {\setlength{\fboxsep}{1pt}\colorbox[HTML]{d8d8d8}{\textit{English}}} translations.
Error spans are marked with {\setlength{\fboxsep}{1pt}\colorbox[HTML]{FF9999}{\text{red}}} highlights to indicate the exact locations of the issues.
}
\label{fig:error_analysis}
\end{figure*}

\subsection{Error Analysis}
\label{subsubsec:error_pattern}

Generating counterfactuals is not immune to errors, possibly due to the suboptimal instruction-following ability of LLMs and their difficulty in handling fine-grained semantic changes.
\citet{nguyen-etal-2024-llms} have identified three common categories of errors in English counterfactuals.
We hypothesize that similar issues may arise in multilingual counterfactual generation.
Building on this insight, we examine the directly generated counterfactuals $\tilde{x}_\ell$ in all target languages, analyzing them both manually and automatically, depending on the type of error.
To facilitate our investigation, we translate the counterfactuals into English when necessary and compare them against their original texts.
Based on this process, we identify four distinct error types, which we summarize below (see Figure~\ref{fig:error_analysis} for examples in each error type).



\paragraph{Copy-Paste.} 
When LLMs are prompted to generate counterfactuals by altering the model-predicted label, they occasionally return the original input unchanged as the counterfactual. 
Figure~\ref{fig:repetitive_rate} shows that the copy-paste rate is considerably higher on \data{SIB200} (average: $6.7\%$) than on \data{XNLI} (average: $2.1\%$).
However, the trend in two datasets is not consistent across languages.
High-resource languages like English and Spanish in \data{SIB200} present higher copy-paste rates.
In contrast, lower-resource languages like Hindi and Swahili in \data{XNLI} are most affected by the copy-paste issue.
A closer inspection suggests that LLMs often struggle to sufficiently revise the input to align with the target label, resulting in incomplete or superficial edits.

\paragraph{Negation.} 
For counterfactual generation, LLMs often attempt to reverse the original meaning by introducing explicit negation while preserving most of the context. 
However, this strategy frequently fails to trigger the intended label change, resulting in semantically ambiguous or label-preserving outputs \cite{wang2025truthtwistoptimalmodel}.
A likely reason is that LLMs may rely on shallow heuristics -- negation being a common surface-level cue for meaning reversal learned during pretraining. 
Especially in languages with simple and explicit negation markers, such as English and German, LLMs tend to perform minimal edits (e.g., adding ``not'') rather than making deeper structural changes required for a true semantic shift.

\paragraph{Inconsistency.} 
Counterfactuals may introduce statements that are logically contradictory or incoherent relative to the original input.
This often results from the model appending or modifying content without fully reconciling the semantic implications of the added text with the existing context.
In such cases, the counterfactual may contain mutually exclusive statements, e.g., simultaneously asserting that an event occurred and that it was prohibited (cf. Figure~\ref{fig:error_analysis}).
These inconsistencies highlight the model's difficulty in preserving global meaning while introducing label-altering edits, particularly when attempting to retain much of the original phrasing.

\paragraph{Language Confusion.}
We further identify the language of directly generated counterfactuals $\tilde{x}$ and examine whether it aligns with the intended target language.\footnote{\url{https://github.com/zafercavdar/fasttext-langdetect}}
Figure~\ref{fig:lang_conf} illustrates the language confusion rate  \cite{marchisio-etal-2024-understanding}
across different languages on \data{XNLI} and \data{SIB200}. 
Overall, counterfactuals in high-resource languages, i.e., German, English, and Spanish, can be generated consistently in the expected target language. 
In contrast, when relatively lower-resource languages, such as Arabic or Swahili, are specified as the target language, LLMs frequently misinterpret the prompts\footnote{More discussion about the selection of languages for prompts can be found in Appendix~\ref{app:cfe}.} or default to generate counterfactuals in the predominant language of English \cite{hwang2025learngloballyspeaklocally}. 

\subsection{Counterfactual Data Augmentation}
\label{subsec:cda}

Table~\ref{tab:cda} reflects that for the base model $\mathcal{M}_{base}$, multilingual CDA generally leads to a substantial improvement in performance compared to cross-lingual CDA across two datasets. This effect is especially compelling for Arabic, with average accuracy gains of $64.45\%$, while for English, the improvement is least observable due to its already satisfactory performance in the cross-lingual setting.\looseness=-1

For \data{XNLI}, cross-lingual CDA enhances model performance only on English, while degrading performance on the other languages. In the context of multilingual CDA, overall, model performance improves across languages other than Swahili. For \data{SIB200}, in the cross-lingual setting, CDA generally has an adverse impact on model performance. Meanwhile, although the generated counterfactuals are more effective and valid (Table~\ref{tab:automatic_evaluation}), in the multilingual setting, augmenting with these counterfactuals only yields reliable gains in English and Spanish, while it even consistently hampers performance for Swahili. This effect is remarkably pronounced when using smaller LLMs.

\input{tables/cda}

The limited performance improvement from augmenting with counterfactuals can be attributed to the imperfection of generated counterfactuals (Figure~\ref{fig:error_analysis}), which stems from both the limited multilingual capabilities of LLMs and suboptimal multilingual counterfactual generation method. We take a close look into how error cases (Figure~\ref{fig:error_analysis}) affect the model performance gains achieved through CDA. Table~\ref{tab:cda_error_analysis} reveals that, after excluding error cases (\textit{copy-paste} and \textit{language confusion}), overall performance improves; however, the magnitude of enhancement varies across languages (Appendix~\ref{app:analysis}). Furthermore, while counterfactuals for \data{SIB200} often succeed in flipping model predictions, they frequently fail to flip the ground-truth labels due to insufficient revision, an essential requirements for CDA, resulting in noisy labels that can even deteriorate performance \cite{zhu-etal-2022-bert, PMID:35254993, wang2025truthtwistoptimalmodel}.\footnote{Further details on CDA, including training-data selection, model training, and additional results evaluated using human-annotated counterfactuals, are offered in Appendix~\ref{app:cda}.} 


\section{Conclusion}
In this work, we first conducted automatic evaluations on directly generated counterfactuals in the target languages and translation-based counterfactuals generated by three LLMs across two datasets covering six languages. Our results show that directly generated counterfactuals in high-resource European languages tend to be more valid and effective. Translation-based counterfactuals yield higher LFR than directly generated ones but at the cost of substantially greater editing effort. Nonetheless, these translated variants still fall short of the original English counterfactuals from which they derive. Second, we revealed that the nature and pattern of edits in English, German, and Spanish counterfactuals are strikingly similar, indicating that cross-lingual perturbations follow common strategies. Third, we cataloged four principal error types that emerge in the generated counterfactuals. Of these, the tendency to copy and paste segments from the source text is by far the most pervasive issue across languages and models. Lastly, we extended our study to CDA. Evaluations across languages show that multilingual CDA outperforms cross-lingual CDA, particularly for low-resource languages. However, given that the multilingual counterfactuals are imperfect, CDA does not reliably improve model performance or robustness. 


\section*{Limitations}
We use multilingual sentence embeddings to assess textual similarity between the original input and its counterfactual (\S\ref{subsec:automatic_evaluation}), following \citet{wang-etal-2024-survey, bhattacharjee2024llmguidedcausalexplainabilityblackbox}. While token-level Levenshtein distance is widely adopted as an alternative \cite{ross-etal-2021-explaining, treviso-etal-2023-crest, wang-etal-2025-fitcf}, it may not fully capture similarity for non-Latin scripts. This underscores the need for new token-level textual similarity metrics suited to multilingual settings.

We do not exhaustively explore all languages common to \data{SIB200} and \data{XNLI}; instead, we select 6 languages spanning from high-resource to low-resource to ensure typological diversity and cover a variety of scripts (\S\ref{subsec:datasets}). Thus, expanding the evaluation to more languages and exploring more models with different architectures and sizes are considered as directions for future work.

Since machine translation quality is not strongly correlated with the improvement of counterfactual validity (\S\ref{subsubsec:translation_results}). Therefore, approaches based on machine translation may not be an optimal method for multilingual counterfactual generation. The quality of multilingual counterfactuals could potentially be considerably improved by adopting post-training methods, such as MAPO \cite{she-etal-2024-mapo}, serving as a promising way for future work. 


In this work, following prior research on comprehensive studies of English counterfactuals \cite{nguyen-etal-2024-llms, wang-etal-2024-survey, mcaleese2024comparativeanalysiscounterfactualexplanation}, we focus exclusively on automatic evaluations of multilingual counterfactuals along three dimensions -- validity, fluency and minimality (\S\ref{subsec:automatic_evaluation}), rather than on subjective aspects such as usefulness, helpfulness, or coherence of counterfactuals \cite{domnich2025unifyingevaluationcounterfactualexplanations, wang2026largelanguagemodelsexplain, wang2026iflipiterativefeedbackdrivencounterfactual}, which can only be assessed through user study. As future work, we plan to conduct a user study to subjectively assess the quality of the multilingual counterfactuals.

\section*{Ethics Statement}
The participants in the machine translation evaluation (Appendix~\ref{subsubsec:user_study}) were compensated at or above the minimum wage,
in accordance with the standards of our host institutions’
regions. The annotation took each annotator approximately an hour on average.

\section*{Author Contributions}
Author contributions are listed according to the CRediT taxonomy as follows:
\begin{itemize}[noitemsep,topsep=0pt,leftmargin=*]
    \item QW: Writing, idea conceptualization, experiments and evaluations, analysis, user study, visualization.
    \item VBN: Writing, preparation and evaluation of human-annotated counterfactuals for multilingual CDA.
    \item YL: Writing and error analysis.
    \item FS: Multilingual CDA on test set.
    \item NF: Writing – review \& editing and supervision.
    \item CS: Supervision and review \& editing.
    \item HS: Supervision and review \& editing.
    \item SM: Supervision and funding acquisition.
    \item VS: Funding acquisition and proof reading.
\end{itemize}

\section*{Acknowledgment}
We thank Jing Yang for her insightful feedback on earlier drafts of this paper, valuable suggestions and in-depth discussions.
We sincerely thank Pia Wenzel, Zain Alabden Hazzouri, Luis Felipe Villa-Arenas, Cristina España i Bonet, Salano Odari, Innocent Okworo, Sandhya Badiger and Juneja Akhil for evaluating translated texts.
Additionally, we are indebted to the anonymous reviewers of EACL 2026 for their helpful and rigorous feedback.
This work has been supported by the Federal Ministry of Research, Technology and Space (BMFTR) as part of the projects BIFOLD 24B and VERANDA (16KIS2047).

\bibliography{custom}

\appendix

\section{Counterfactual Generation}
\label{app:cfe}

Figure~\ref{fig:cf_prompt_xnli} and Figure~\ref{fig:cf_prompt_sib} demonstrate prompt instructions for counterfactual example generation on the \data{XNLI} and \data{SIB200} datasets. An example from each dataset is included in the prompt. Figure~\ref{fig:translation} illustrates the prompt instruction for translating a counterfactual example from English to a target language.

\begin{figure*}[h!]
    
    \centering

    \begin{tcolorbox}[colback=blue!30!white, colframe=black!10!blue, title=XNLI (Natural Language Inference)]

    Given two sentences (premise and hypothesis) in \{language\_dict[language]\} and their original relationship, determine whether they
    entail, contradict, or are neutral to each other. Change the premise with minimal edits to achieve
    the target relation from the original one and output the edited premise surrounding by <edit>[premise]</edit> in {language}. 
    Do not make any unnecessary changes.
    \bigskip
    
    \#\#\#\#\#Begin Example\#\#\#\#
    
    \textbf{Original relation:} \textit{entailment}
    
    \textbf{Premise:} A woman is talking to a man. 
    
    \textbf{Hypothesis:} Brown-haired woman talking to man with backpack.
    
    \textbf{Target relation:} \textit{neutral}

    \bigskip
    
    \textbf{Step 1:} Identify phrases, words in the premise leading to the entailment relation:
    'man';
    
    \textbf{Step 2:} Change these phrases, words to get neutral relation with minimal changes:
    'man' to 'student';
    
    \textbf{Step 3:} replace the phrases, words from step 1 in the original text by the phrases, words, sentences
    in step 2.
    \bigskip
    
    \textbf{Edited premise:} <edit>A woman is talking to a student.</edit>
    
    \#\#\#\#\#End Example\#\#\#\#\#

    \bigskip

    \textbf{Request:} Given two sentences (premise and hypothesis) in \{language\_dict[language]\} and their original relationship, determine
    whether they \textit{entail}, \textit{contradict}, or are \textit{neutral} to each other. Change the \textbf{premise} with minimal edits
    to achieve the neutral relation from the original one  and output the edited premise surrounding by 
    <edit>[premise]</edit> in \{language\_dict[language]\}. Do not make any unnecessary changes. Do not add anything else.
    \bigskip

    \textbf{Original relation:} \{prediction\}
    
    \textbf{Premise}: \{premise\}
    
    \textbf{Hypothesis}: \{hypothesis\}
    
    \textbf{Target relation:} \{target\_label\}
    
    \textbf{Edited premise:}

    \end{tcolorbox}

    \caption{Prompt instruction for counterfactual example generation on the \data{XNLI} dataset.
    }
    \label{fig:cf_prompt_xnli}
\end{figure*}

\begin{figure*}[h!]
    
    \centering
    \begin{tcolorbox}[colback=blue!30!white, colframe=black!10!blue, title=SIB200 (Topic Classification)]

    Given a sentence in \{language\_dict[language]\} classified as belonging to one of the topics: 
``science/technology'', ``travel'', ``politics'', ``sports'', ``health'', ``entertainment'', ``geography''. Modify the 
sentence to change its topic to the specified target topic and output the edited sentence surrounding by 
<edit>[sentence]</edit> in {language}.
Do not make any unnecessary changes.

\bigskip
\#\#\#\#\#Begin Example\#\#\#\#\#

\textbf{Original topic:} \textit{sports}

\textbf{Sentence}: The athlete set a new record in the marathon.

\textbf{Target topic:} \textit{health}
\bigskip

\textbf{Step 1:} Identify key phrases or words determining the original topic:
'athlete', 'record', 'marathon'.

\textbf{Step 2:} Modify these key phrases or words minimally to reflect the target topic (health):
'athlete' to 'patient', 'set a new record' to 'showed improvement', 'marathon' to 'rehabilitation'.

\textbf{Step 3:} Replace the identified words or phrases in the original sentence:
\bigskip

\textbf{Edited sentence:} <edit>The patient showed improvement in the rehabilitation.</edit>

\#\#\#\#\#End Example\#\#\#\#\#
\bigskip

\textbf{Request:} Given a sentence in \{language\_dict[language]\} classified as belonging to one of the topics: 
``science/technology'', ``travel'', ``politics'', ``sports'', ``health'', ``entertainment'', ``geography''. Modify the 
sentence to change its topic to the specified target topic and output the edited sentence surrounding by 
<edit>[sentence]</edit> in {language}.
Do not make any unnecessary changes.
\bigskip

\textbf{Original topic:} \{prediction\}

\textbf{Sentence}: \{text\}

\textbf{Target topic: } \{target\_label\}

\textbf{Edited sentence:}
        
    \end{tcolorbox}
    \caption{Prompt instruction for counterfactual example generation on the \data{SIB200} dataset.
    }
    \label{fig:cf_prompt_sib}
\end{figure*}

\begin{figure*}[h!]
    
    \centering

    \begin{tcolorbox}[colback=blue!30!white, colframe=black!10!blue, title=Machine Translation]

You are a professional translator, fluent in English and \{language\}. Translate the following English text to \{language\} accurately and naturally, preserving its tone, style, and any cultural nuances.
 \bigskip
Text to translate: \{counterfactual\}

    \end{tcolorbox}

    \caption{Prompt instruction for translating a counterfactual example from English to a target language.
    }
    \label{fig:translation}
\end{figure*}

\section{Datasets}
\label{app:dataset}

\subsection{Dataset Examples}
\begin{figure*}[h]
\centering
\resizebox{0.85\textwidth}{!}{
\begin{minipage}{\columnwidth}
\includegraphics[width=\columnwidth]{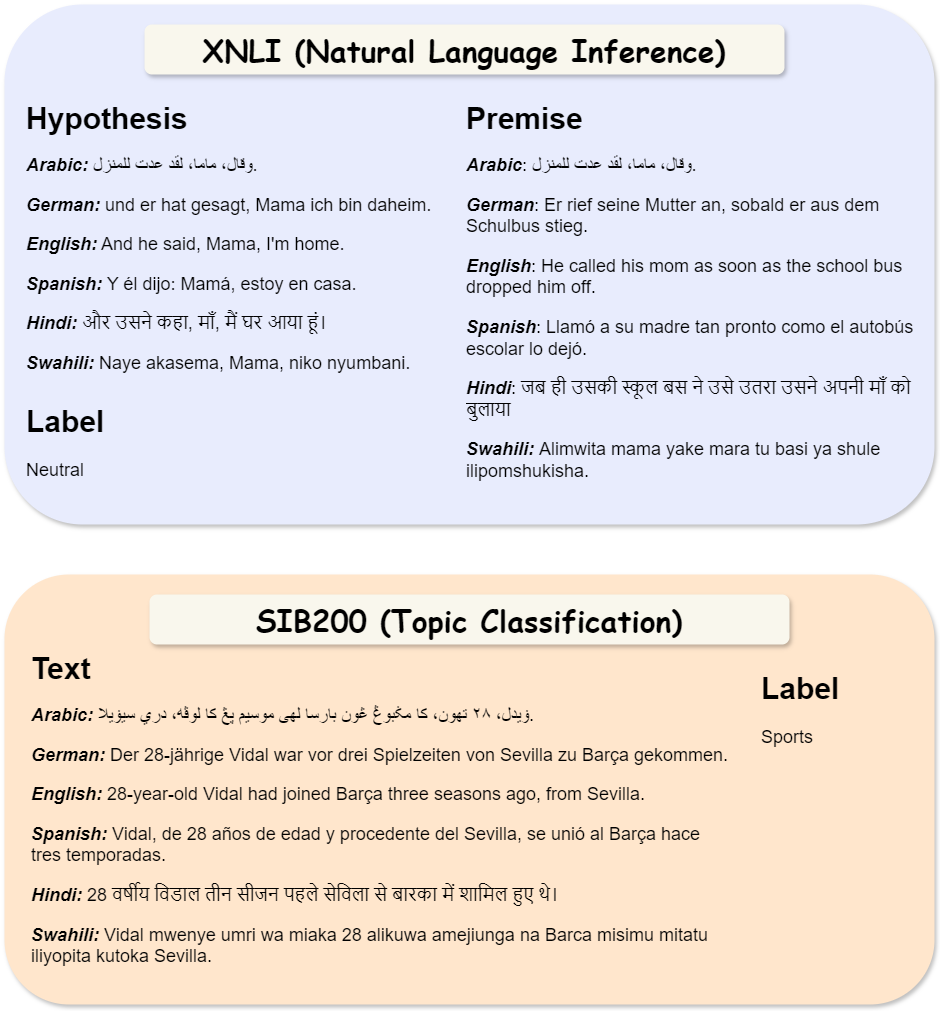}
\end{minipage}
}
\caption{Examples from the \data{XNLI} and \data{SIB200} dataset.}
\label{fig:xnli_example}
\end{figure*}

Figure~\ref{fig:xnli_example} presents parallel examples from the \data{XNLI} and \data{SIB200} datasets in \textit{Arabic}, \textit{German}, \textit{English}, \textit{Spanish}, \textit{Hindi} and \textit{Swahili}. 

\subsection{Label Distributions}
\begin{figure}[t!]
  \centering

  \begin{subfigure}{\columnwidth}
    \centering
    \centering
\includegraphics[width=\linewidth]{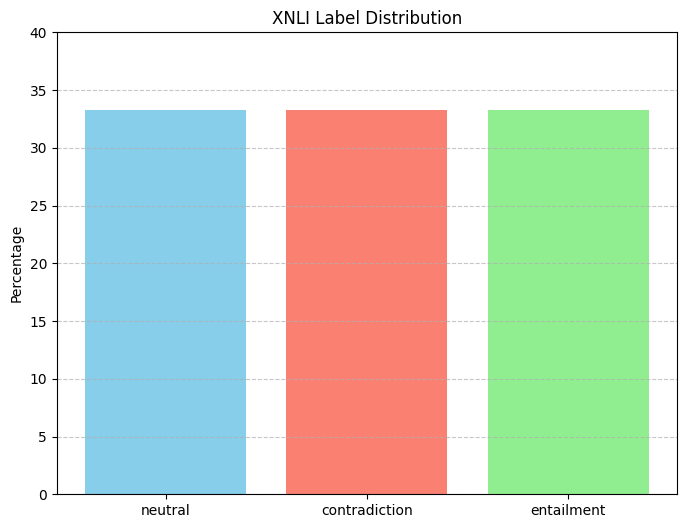}
\caption{\data{XNLI}}
\label{fig:xnli_distribution}
  \end{subfigure}
  \hfill
  \begin{subfigure}{\columnwidth}
\centering
\includegraphics[width=\linewidth]{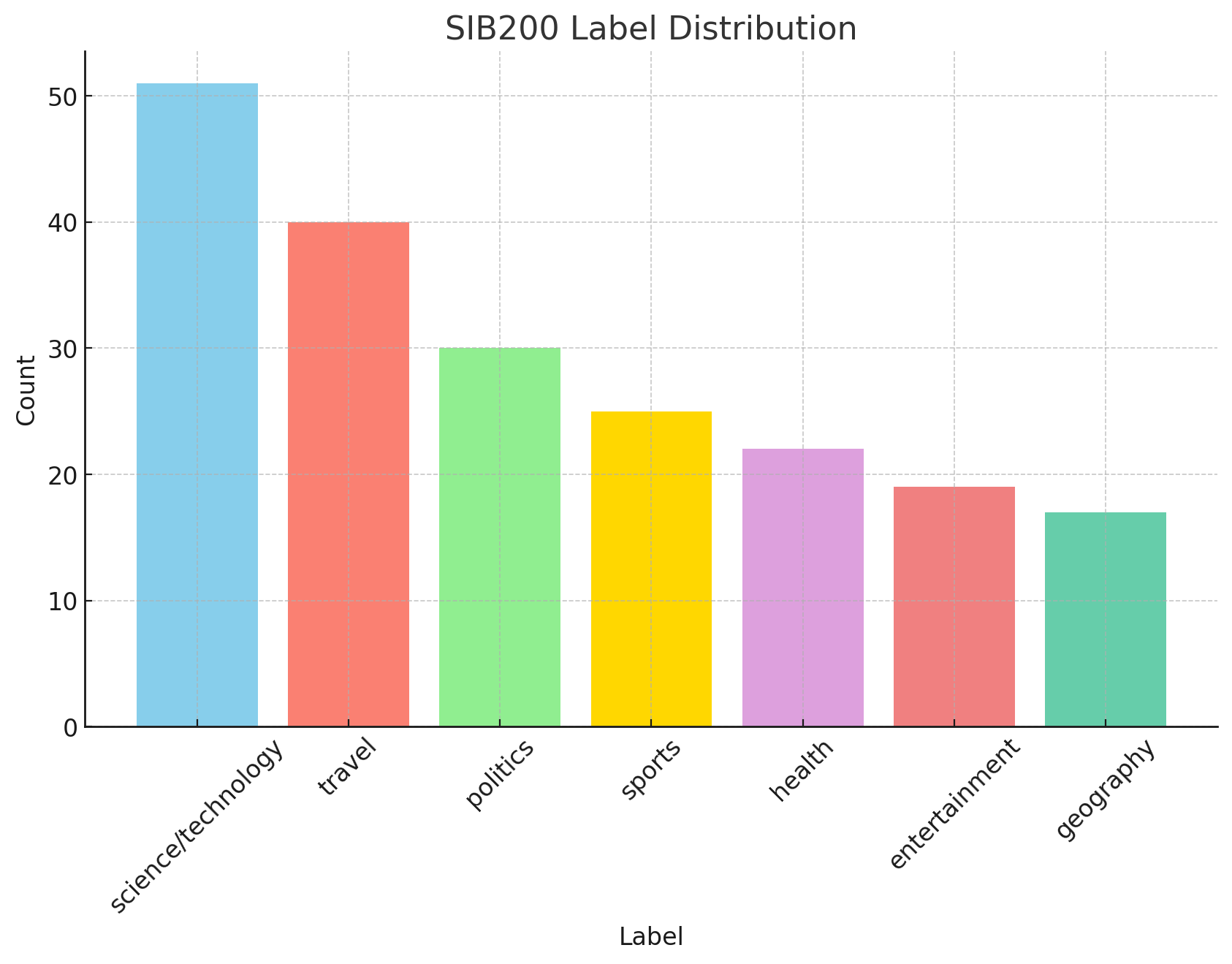}
\caption{\data{SIB200}}
\label{fig:sib_label_distribution}

  \end{subfigure}

  \caption{Label distributions of \data{XNLI} and \data{SIB200}.}
  \label{fig:app_label_distribution}
\end{figure}

Figure~\ref{fig:app_label_distribution} illustrates the label distributions for \data{XNLI} and \data{SIB200}.


\section{Experiment}
\label{app:experiment}

\subsection{Models}

\input{tables/models}
\input{tables/language}

\subsubsection{LLMs for Counterfactual Generation}
\label{app:model}
Table~\ref{tab:used_model} displays three open-source LLMs are utilized for counterfactual generation (\S\ref{subsec:models}).

Table~\ref{tab:language} shows language support for the selected languages as shown in \S\ref{subsec:datasets}. For \lm{Qwen2.5-7B}, the model supports additional languages beyond those listed in Table~\ref{tab:language}; however, these are not specified in the technical report \cite{qwen2024qwen25technicalreport}. Similarly, \lm{Gemma3-27B} is reported to support over 140 languages \cite{gemmateam2025gemma3technicalreport}, though the exact supported languages are not disclosed.

\subsubsection{Explained Models}
\label{app:explained_model}
\input{tables/sib200_performance}
Table~\ref{tab:sib200_performance} presents the task performance of the explained models $\mathcal{M}_{ft}$ (\S\ref{subsec:automatic_evaluation}) across all identified languages on the \data{XNLI} and \data{SIB200} datasets. For \data{XNLI}, we use the fine-tuned \lm{mBERT} model, which is publicly available and downloadable directly from Huggingface\footnote{\url{https://huggingface.co/MayaGalvez/bert-base-multilingual-cased-finetuned-nli}}. For \data{SIB200}, we fine-tuned a pretrained \lm{mBERT} on the \data{SIB200} training set.

\paragraph{\lm{mBERT} fine-tuning on \data{SIB200}} We fine-tuned \lm{bert-base-multilingual-cased}\footnote{\url{https://huggingface.co/google-bert/bert-base-multilingual-cased}} for 7-way topic classification (Figure~\ref{fig:sib_label_distribution}). The input CSV contains a text column with multilingual content stored as a Python dict (language$\rightarrow$text) and a categorical label. Each row is expanded so that every language variant becomes its own training example while inheriting the same label. We split the expanded dataset into 80\% train / 20\% validation with a fixed random seed. We train with the Hugging Face Trainer\footnote{\url{https://huggingface.co/docs/transformers/main_classes/trainer}} using linear LR schedule with 500 warmup steps, for 3 epochs, at a learning rate $2e^{-5}$, with a batch size 16 and weight decay of 0.01. We evaluate once per epoch and save a checkpoint at the end of each epoch. The best checkpoint is selected by macro-$F_1$ and restored at the end. Early stopping monitors macro-$F_1$ with a patience of one evaluation round.

\input{tables/inference}

\subsection{Inference Time}

Table~\ref{tab:inference} displays inference time for counterfactual generation per language using \lm{Qwen2.5-7B}, \lm{Gemma3-27B}, and \lm{Llama3.3-70B} on \data{XNLI} and \data{SIB200}.

\section{Machine Translation Evaluation}
\label{app:translation}
\subsection{Automatic Evaluation}
\label{subsubsec:automatic_evaluation}
Given that we explore translation-based counterfactuals (\S\ref{subsec:cfe_generation}), we employ three commonly used automatic evaluation metrics to assess translation quality at different levels of granularity, following \citet{zhang-etal-2023-translation, pang-etal-2025-salute, pei-etal-2025-understanding}. 

\paragraph{BLEU} \cite{papineni-etal-2002-bleu} measures how many \textbf{n-grams} (contiguous sequences of words) in the candidate translation appear in the reference.

\paragraph{chrF} \cite{popovic-2015-chrf} measures overlap at the \textbf{character n-gram level} and combines precision and recall into a single F-score, better capturing minor orthographic and morphological variations.


\paragraph{XCOMET} \cite{guerreiro-etal-2024-xcomet} is a learned metric that simultaneously perform \textbf{sentence-level} evaluation and error span detection. In addition to providing a single overall score for a translation, XCOMET highlights and categorizes specific errors along with their severity.

All three selected metrics are reference-based. However, since we do not have ground-truth references (i.e., gold-standard counterfactuals in the target languages), we perform back-translation \cite{sennrich-etal-2016-improving} by translating the LLM-translated counterfactuals $\tilde{x}_{\textsf{en}\text{-}\ell}$ (\S\ref{subsec:cfe_generation}) back into English, yielding $\tilde{x}_{\text{back}}$. We then compare $\tilde{x}_{\text{back}}$ with the original English counterfactuals $\tilde{x}_{\textsf{en}}$ (Table~\ref{tab:automatic_evaluation_translation}), known as round-trip translation \cite{somers-2005-round, moon-etal-2020-revisiting, zhuo-etal-2023-rethinking}.

\subsection{Human Evaluation}
\label{subsubsec:user_study}
To further validate the multilingual counterfactual examples translated by LLMs $\tilde{x}_{\textsf{en}\text{-}\ell}$ (\S\ref{subsec:cfe_generation}) beyond automatic evaluation metrics, we conducted a human evaluation in the form of Direct Assessment (DA) \cite{graham-etal-2013-continuous} on a continuous scale from 0 to 100, following \citet{pei-etal-2025-understanding}. Note that in this user study, we only evaluate the quality of machine translated texts instead of assessing the quality of multilingual counterfactual explanations. We randomly select ($k=10$) dataset indices for \data{XNLI} and \data{SIB200}. For each subset, i.e., model-language pair (Table~\ref{tab:automatic_evaluation}), the translated counterfactuals in the target language, generated by the given model for the selected indices, are evaluated by two human annotators. The counterfactuals are presented to annotators in the form of questionnaires. We recruit $n=10$ in-house annotators, all of whom are native speakers of one of the selected languages (\S\ref{subsec:datasets}). Figure~\ref{fig:annotation_guideline} illustrates the annotation guidelines provided to human annotators for evaluating the quality of machine translation texts. 


\begin{figure*}[h!]
    
    \centering

    \begin{tcolorbox}[colback=orange!20!white, colframe=black!20!orange, title=Annotation Guideline]
    Dear Participants,
\bigskip

Thank you for being part of the annotation team. in this study, we will evaluate machine-translated texts generated by large language models (LLMs) from English source texts. You will be presented with pairs of texts: the original text in English and its translation in one of the following target languages—German, Spanish, Arabic, Hindi, or Swahili.

\bigskip
Your evaluation should be based on how well the translation captures the meaning of the sentence in the target language. The English reference translation is included only to help resolve ambiguities — do not score based on how closely the system-generated translation matches the reference. If the system-generated translation uses different words or phrasing but adequately conveys the meaning, it should not be penalized.

\bigskip

The focus of this evaluation is on how well the system-generated English translation conveys the meaning of the sentence in the target language. Do not penalize translations for awkward or unnatural English phrasing as long as the meaning is adequately preserved. You will use a slider (0–100) to score each translation.

• 0\% – No meaning preserved

• 33\% – Some meaning preserved

• 66\% – Most meaning preserved

• 100\% – Adequate translation, all meaning preserved

\bigskip
Please try to be consistent in your use of the scale across all items. Your valuable evaluation willhelp improve the quality of machine translation for endangered languages.

    \end{tcolorbox}
    
    \caption{Annotation guideline provided to human annotators for evaluating the quality of machine translation texts.
    }
    \label{fig:annotation_guideline}
\end{figure*}

\subsection{Results}
\label{subsubsec:translation}
\subsubsection{Automatic Evaluation}
\input{tables/automatic_translation}

Table~\ref{tab:automatic_evaluation_translation} displays that, overall, Spanish and German translations exhibit higher quality compared to Arabic, Hindi, and Swahili across various evaluation metrics with different levels of granularity (\S\ref{subsubsec:automatic_evaluation}). We observe a strong correlation between BLEU and XCOMET, with Spearman's $\rho$ of 0.89 for \data{XNLI} and 0.77 for \data{SIB200}.


\subsubsection{Human Evaluation}
\label{subsec:human_evaluation}

\input{tables/judge_avg}
\input{tables/correlation}
\input{tables/translation_analysis}


Table~\ref{tab:human_evaluation} delivers direct-assessment (DA scores for back-translated counterfactuals $\tilde{x}_{\textsf{en}\text{-}\ell}$ on \data{XNLI} and \data{SIB200}. Overall, Arabic, Spanish, and German back-translations achieve good quality. Notably, \lm{Qwen2.5-7B} exhibits markedly poorer Swahili translation quality than the other two models.

\paragraph{Correlation with Automatic Metrics.} Table~\ref{tab:correlation} illustrates Spearman's rank correlation ($\rho$) between automatic evaluation metric results and human evaluation results. We observe that BLEU and XCOMET show moderate correlations with human judgments, whereas chrF correlates positively on \data{XNLI} but negatively on \data{SIB200}.

\paragraph{Agreement.} Table~\ref{tab:translation_analysis} reports inter-annotator agreement (IAA) scores and associated $p$-values for all languages (\S\ref{subsec:datasets}) except English. Annotators show high agreement for Swahili, whereas German exhibits comparatively low agreement. Nevertheless, the $p$-values indicate that the observed agreements are statistically significant.

\section{Evaluation}

\subsection{Perplexity}
\input{tables/perplexity}
Table~\ref{tab:perplexity} illustrates the perplexity scores of data points across the selected languages (\S\ref{subsec:datasets}) from the \data{XNLI} and \data{SIB200} datasets. We observe that on \data{XNLI}, the \textit{Hindi} premises and hypotheses exhibit the highest fluency, whereas the \textit{English} ones exhibit the lowest. On \data{SIB200}, the \textit{German} texts are the most fluent, while the \textit{Arabic} texts are the least fluent.

\subsection{Cross-lingual Edit Similarity}
\label{app:crosslingual_sim}

\subsubsection{Cosine Similarity of Original Inputs}
\begin{figure}[t!]
  \centering

  \begin{subfigure}{\columnwidth}
    \centering
    \centering
\includegraphics[width=\linewidth]{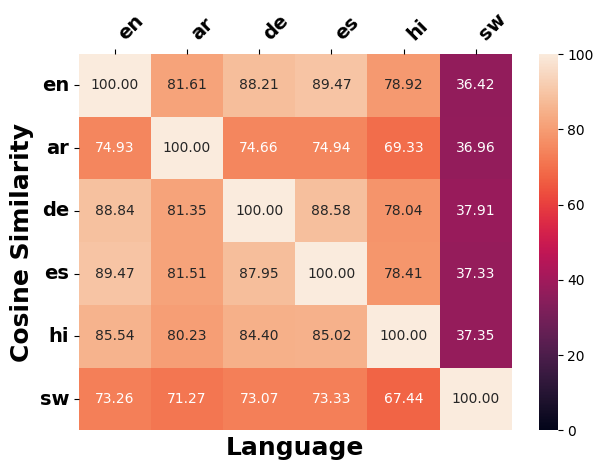}
\caption{\data{XNLI}}
\label{fig:cos_sim_XNLI_dataset}
  \end{subfigure}
  \hfill
  \begin{subfigure}{\columnwidth}
\centering
\includegraphics[width=\linewidth]{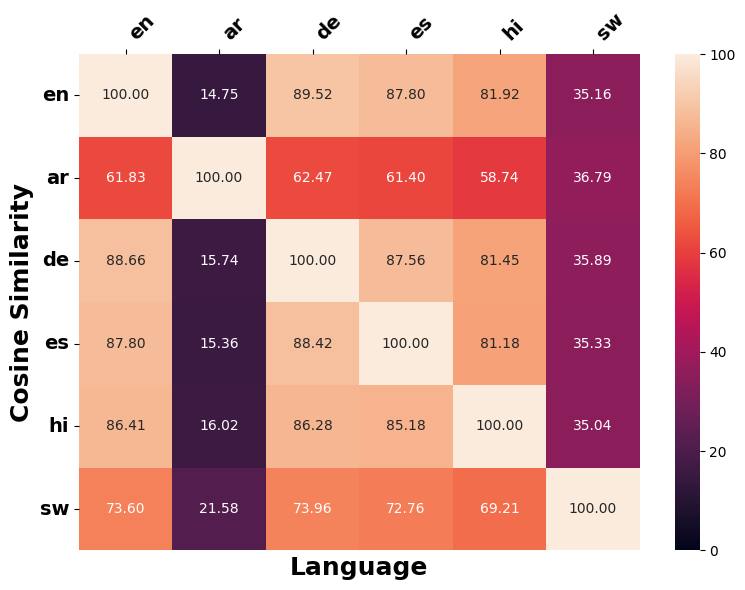}
\caption{\data{SIB200}}
\label{fig:cos_sim_sib200_dataset}

  \end{subfigure}

  \caption{Cosine similarity scores for original inputs across different language from \data{XNLI} and \data{SIB200}.}
  \label{fig:app_cos_sim}
\end{figure}

Figure~\ref{fig:app_cos_sim} illustrates cosine similarity scores for instances across different language from \data{XNLI} and \data{SIB200}. We observe that, despite the availability of parallel data from \data{XNLI} and \data{SIB200}, Swahili texts are generally less similar to those in other languages.

\subsubsection{Cosine Similarity of Back-translated Counterfactuals}
Figure~\ref{fig:app_cos_sim_trans} shows cosine similarity scores for translated counterfactuals $\tilde{x}_{\ell\text{-}\textsf{en}}$ in English across different language $\ell$ from \data{XNLI} and \data{SIB200}. Notably, the translated counterfactuals exhibit significantly lower pairwise similarity compared to the multilingual counterfactuals generated prior to translation.

\subsubsection{Cross-lingual Counterfactual Examples}
\label{app:crosslingual_example}
\begin{figure}[t!]
  \centering

  \begin{subfigure}{\columnwidth}
    \centering
    \centering
\includegraphics[width=\linewidth]{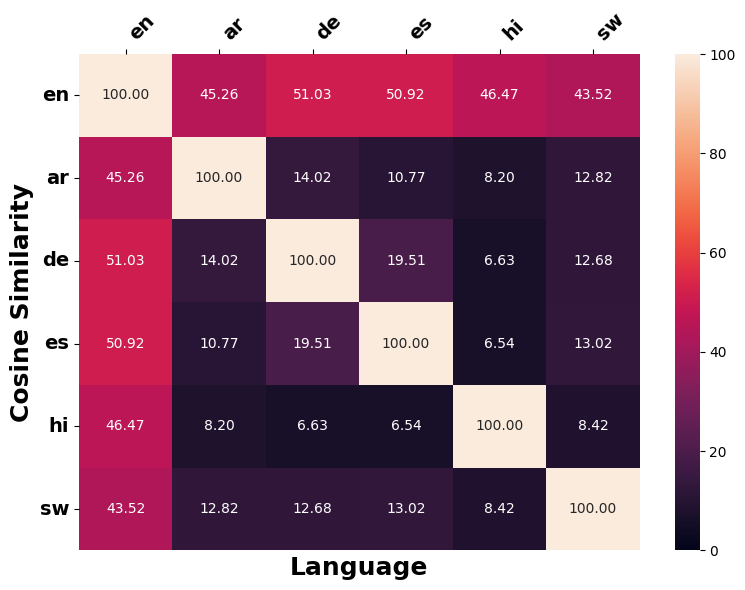}
\caption{\data{XNLI}}
\label{fig:cos_sim_XNLI_trans}
  \end{subfigure}
  \hfill
  \begin{subfigure}{\columnwidth}
\centering
\includegraphics[width=\linewidth]{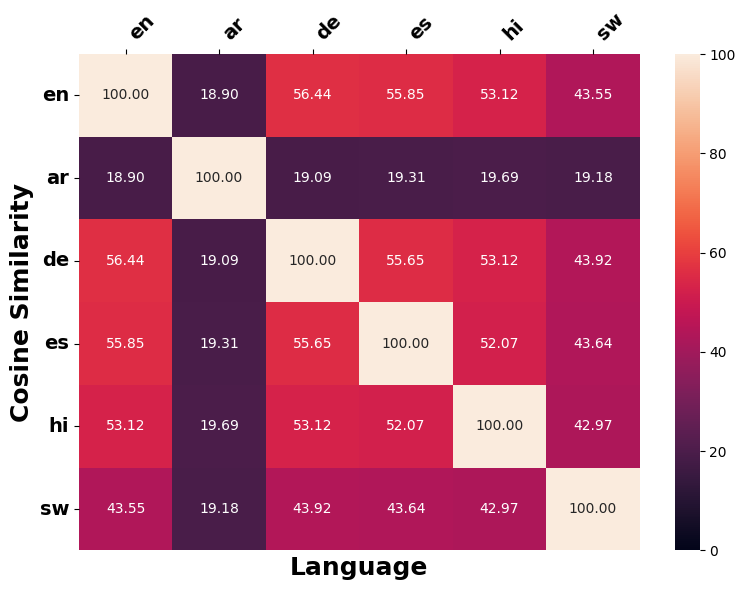}
\caption{\data{SIB200}}
\label{fig:cos_sim_sib200_trans}

  \end{subfigure}

  \caption{Cosine similarity scores for translated counterfactuals in English $\tilde{x}_{\ell\text{-}\textsf{en}}$ across different language $\ell$ from \data{XNLI} and \data{SIB200}.}
  \label{fig:app_cos_sim_trans}
\end{figure}

\begin{figure*}[t]
\centering
\resizebox{\textwidth}{!}{
\begin{minipage}{\columnwidth}
\includegraphics[width=\columnwidth]{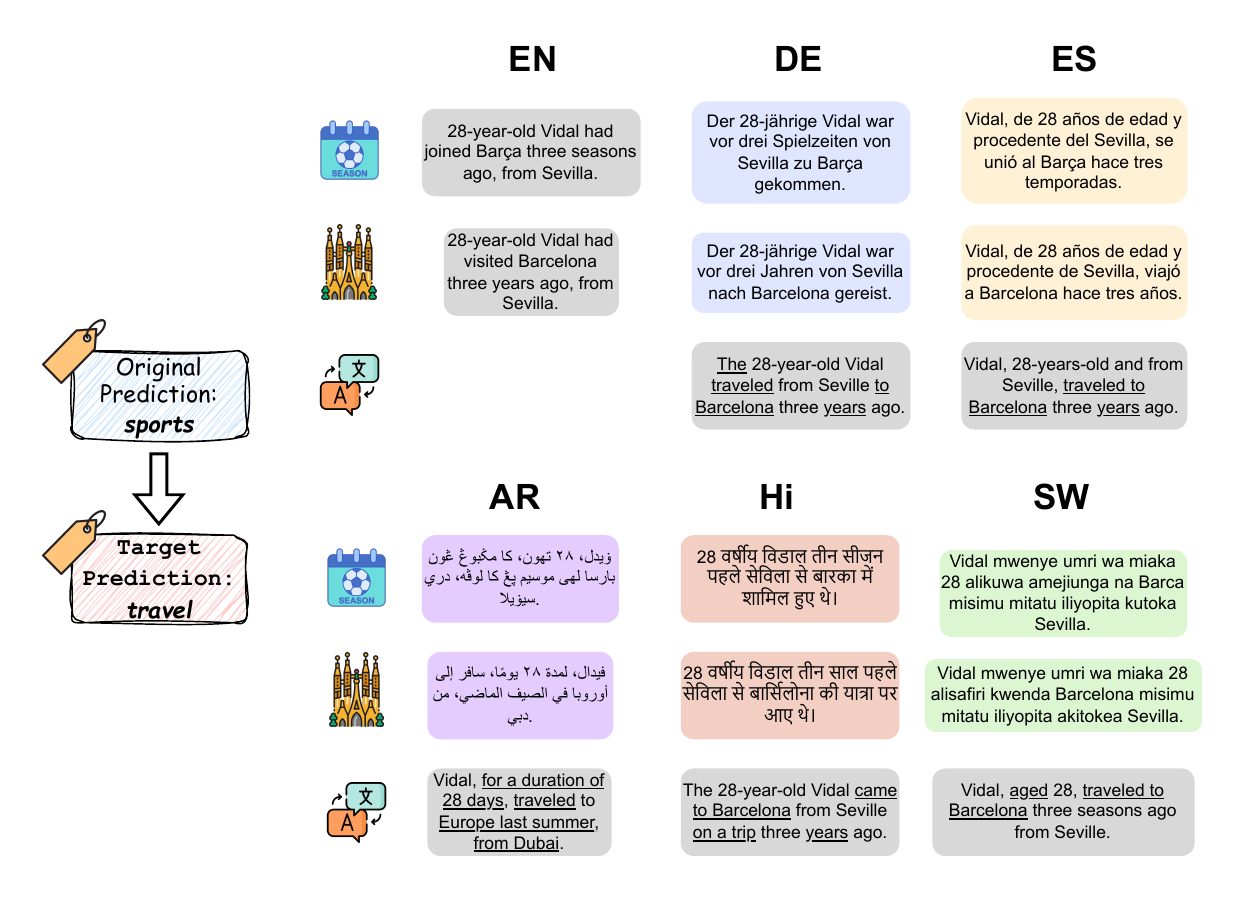}
\end{minipage}
}
\caption{
Original texts, counterfactuals in {\setlength{\fboxsep}{1pt}\colorbox[HTML]{d8d8d8}{\textit{English}}}, 
{\setlength{\fboxsep}{1pt}\colorbox[HTML]{E5CCFF}{\textit{Arabic}}}, {\setlength{\fboxsep}{1pt}\colorbox[HTML]{E0E6FF}{\textit{German}}}, {\setlength{\fboxsep}{1pt}\colorbox[HTML]{FFF1D5}{\textit{Spanish}}}, {\setlength{\fboxsep}{1pt}\colorbox[HTML]{F2CFC2}{\textit{Hindi}}}, and {\setlength{\fboxsep}{1pt}\colorbox[HTML]{DDF6D2}{\textit{Swahili}}} (changing the label from \textbf{``sports''} to \textbf{``travel''}), and their corresponding English translations. Edited spans are \underline{underlined}.
}
\label{fig:crosslingual_example}
\end{figure*}

To further probe cross-lingual edit behavior beyond pairwise cosine similarity, we qualitatively examine how LLMs modify the original inputs across languages. Figure~\ref{fig:crosslingual_example} presents counterfactuals in all selected languages that aim to change the label from \textbf{sports} to \textbf{travel}. Consistent with Figure~\ref{fig:app_cos_sim_trans}, European languages (English, German, Spanish) show largely parallel edit strategies during counterfactual generation. These modifications underlined in Figure~\ref{fig:crosslingual_example} reveal lexical and structural convergence when LLMs edit the original input for counterfactual generation and \textbf{verbs} and \textbf{nouns} are replaced with similar words in most cases (e.g., replacing ``join'' with ``travel'' or ``visit'' and ``season'' with ``year''). 

By contrast, the Arabic example employs a markedly different strategy and, in this instance, introduces geographic bias via the insertion of ``Dubai''. For Swahili, the model often fails to fully alter the original semantic -- e.g., retaining ``three reasons'', which should be replaced to remove sport-specific content -- resulting in ambiguous labels.  

\subsection{Counterfactual Data Augmentation}
\label{app:cda}



\subsubsection{Training Data for CDA}

For models fine-tuned on \data{XNLI}, our training data is randomly sampled from the validation split, while evaluation is conducted on the test split. For \data{SIB200}, our training data is randomly sampled from the training split, while evaluation uses the development split. The respective splits were chosen because of their limited sizes.

Counterfactual instances are loaded from pre-computed files, with each counterfactual example paired with its predicted label as determined by the generating LLM. For $\mathcal{M}_{base}$, models are trained exclusively on original examples with their ground-truth labels. For CDA, the training data is augmented by including both original instances and their corresponding counterfactual variants with their predicted labels, effectively doubling the dataset size.

\subsubsection{Model Training Details}

All CDA models are based on \texttt{bert-base-multilingual-cased} \cite{devlin-etal-2019-bert} and fine-tuned for sequence classification using AdamW optimizer with cosine learning rate scheduling, 0.1 warmup ratio, 0.01 weight decay, 4 gradient accumulation steps, and random seed 42. Training parameters are optimized separately for each dataset and counterfactual generation model through grid search.

\paragraph{Dataset-Specific Configurations}
\data{XNLI} models use larger training sets with shorter sequences, while \data{SIB200} models employ smaller training sets with longer training schedules. Maximum training set sizes are constrained by dataset and split selection: 2,400 examples for \data{XNLI} (validation split) and 700 examples for \data{SIB200} (training split). Training sizes within these limits vary across models due to grid search optimization.

\paragraph{Counterfactual Model Variations}
For \data{SIB200}, all best performing models use identical parameters regardless of counterfactual generation model or cross-lingual vs. multilingual configuration: 700 training examples, 20 epochs, batch size 8, maximum sequence length 192, and learning rate 8e-06. 
For \data{XNLI}, models trained with counterfactuals generated by different LLMs exhibit distinct hyperparameter configurations in our grid search, except for a shared maximum sequence length of 256. $\mathcal{M}_c$ and $\mathcal{M}_m$ augmented by counterfactuals generated by \lm{Gemma3-27B} use identical parameters compared to baseline models, while models trained with counterfactuals generated by \lm{Qwen2.5-7B} and \lm{Llama3.3-70B} use different learning rates, batch sizes, and training schedules in the explored parameter space, as shown in Table~\ref{tab:model_training_params}.

\input{tables/model_training_params}

\subsubsection{Human Annotated Counterfactuals}
\label{app:human_annotated_cfs}
Apart from evaluating base models and counterfactually augmented models on the test set from the original datasets, we also prepare human-annotated counterfactuals, which can be considered as out-of-distribution data. For \data{XNLI}, we extend the English counterfactuals from \data{SNLI} \cite{bowman-etal-2015-large} provided by \citet{Kaushik2020Learning}\footnote{\url{https://github.com/acmi-lab/counterfactually-augmented-data}} and translate them into target languages with \lm{Llama3.3-70B}\footnote{We argue that the translation quality should be similar to that shown in Table~\ref{tab:automatic_evaluation_translation} and Table~\ref{tab:human_evaluation}, since we use the same \lm{Llama3.3-70B} model, and thus we leave the machine translation evaluation out.} with the same prompt used in Figure~\ref{fig:translation}. For \data{SIB200}, we ask our in-house annotators to manually create the English counterfactuals. For those, we keep the target label distribution as balanced as possible to avoid any label biases. Similarly, we translate them into target languages with \lm{Llama3.3-70B}.

\subsubsection{Results}
\paragraph{Directly Generated Counterfactual Data Augmentation.} Table~\ref{tab:cda_human}
\input{tables/cda_human} displays the CDA results on human-annotated counterfactuals (\S\ref{app:human_annotated_cfs}). Aligned with the findings on the original dataset (\S\ref{subsec:cda}), multilingual CDA simultaneously yields greater robustness gains, evidenced by higher accuracy, than cross-lingual CDA. On \data{SIB200}, the robustness of counterfactually data augmented models generally improves across all languages, with occasional declines in Hindi and Swahili. The gains are more pronounced for cross-lingual CDA, particularly for English, Spanish, and German. For \data{XNLI}, CDA reduces model robustness, with a consistent degradation observed on the English subset, whereas for Arabic, Hindi, and Swahili, multilingual CDA results in noticeable robustness enhancements.

\input{tables/cda_translation}

\paragraph{Translation-based Counterfactual Data Augmentation.} Since cross-lingual CDA includes only English counterfactuals, we omit these results in Table~\ref{tab:cda_translation}, as they are identical to Table~\ref{tab:cda} and Table~\ref{tab:cda_human}. Table~\ref{tab:cda_translation} shows that for translation-based counterfactual data augmentation, multilingual CDA yields noticeably better model performance than cross-lingual CDA, particularly for lower-resource languages (Arabic, Hindi and Swahili) -- a pattern consistent with our findings for directly generated counterfactual augmentation. Specifically, the cross-lingual CDA generally hampers model robustness, with exceptions for Arabic on \data{XNLI} and English on \data{SIB200}.

\subsubsection{Error Analysis}
\label{app:analysis}
\input{tables/cda_error_analysis}
We provide additional evidence showing how error cases affect the model performance enhancement achieved through counterfactual data augmentation. While copy-paste and language confusion cases are easily detectable using tools or regular expressions, the manual recognition of inconsistency and negation is highly time-consuming. We, therefore, conducted a small-scale CDA experiment (on \data{XNLI} with counterfactuals generated by \lm{Qwen2.5-7B}) that specifically filtered out these easily detectable cases.

Table~\ref{tab:cda_error_analysis} reveals that after filtering out error cases (\textit{copy-paste} and \textit{language confusion}), model performance is improved across all languages. The improvement on English is limited, since the error cases in English are rather rare. The extent of this improvement is directly related to the percentage of initial error cases. For instance, Hindi and Swahili exhibited higher rates of both copy-paste and language confusion (Figure~\ref{fig:error_analysis}); consequently, after filtering, these languages achieved greater performance gains compared to English or other high-resource European languages.

\end{document}

%% file: tables/automatic_evaluation.tex
\begin{table*}[t!]
    \centering
    \setlength{\extrarowheight}{3pt}
    \renewcommand*{\arraystretch}{1}
    \footnotesize
    
    \begin{tabular}{c@{\hspace{1em}}c}
    \begin{subtable}[c]{0.48\textwidth}
      \centering
      \resizebox{\textwidth}{!}{%
    \begin{tabular}{ccccc|ccc}
        \toprule[1.5pt]
         \multirow{2}{*}{\rotatebox[origin=c]{90}{\textbf{\scriptsize{Model}}}} & \textbf{Lang-} & \multicolumn{3}{c|}{\textbf{XNLI}} & \multicolumn{3}{c}{\textbf{SIB200}}\\ 
         
          & \textbf{uage} & LFR $\uparrow$ & PPL $\downarrow$ & TS $\uparrow$ & LFR $\uparrow$ & PPL $\downarrow$ & TS $\uparrow$\\
          
          \midrule

        \multirow{6}{*}{\rotatebox[origin=c]{90}{\scriptsize{\lm{Qwen2.5-7B}}}} 
          & \cellcolor[HTML]{d8d8d8}\textsf{en} & \cellcolor[HTML]{d8d8d8}45.42\% & \cellcolor[HTML]{d8d8d8}36.68 & \cellcolor[HTML]{d8d8d8}\underline{0.8818} & \cellcolor[HTML]{d8d8d8}\textbf{92.16\%} & \cellcolor[HTML]{d8d8d8}46.30 & \cellcolor[HTML]{d8d8d8}0.8483\\
          
          & \cellcolor[HTML]{E5CCFF}\textsf{ar} & \cellcolor[HTML]{E5CCFF}44.10\% & \cellcolor[HTML]{E5CCFF}\underline{36.75} & \cellcolor[HTML]{E5CCFF}0.8853& \cellcolor[HTML]{E5CCFF}89.22\% & \cellcolor[HTML]{E5CCFF}\underline{124.37} & \cellcolor[HTML]{E5CCFF}\underline{0.6941}\\

          & \cellcolor[HTML]{E0E6FF}\textsf{de} & \cellcolor[HTML]{E0E6FF}46.63\% & \cellcolor[HTML]{E0E6FF}32.85 & \cellcolor[HTML]{E0E6FF}0.8891 & \cellcolor[HTML]{E0E6FF}77.45\% & \cellcolor[HTML]{E0E6FF}34.42 & \cellcolor[HTML]{E0E6FF}0.8157\\

          & \cellcolor[HTML]{FFF1D5}\textsf{es} & \cellcolor[HTML]{FFF1D5}\textbf{49.44\%} & \cellcolor[HTML]{FFF1D5}30.36 & \cellcolor[HTML]{FFF1D5}0.8900 & \cellcolor[HTML]{FFF1D5}\underline{72.55\%} & \cellcolor[HTML]{FFF1D5}26.97 & \cellcolor[HTML]{FFF1D5}0.8152\\

          & \cellcolor[HTML]{F2CFC2}\textsf{hi} & \cellcolor[HTML]{F2CFC2}39.92\% & \cellcolor[HTML]{F2CFC2}\textbf{8.12} & \cellcolor[HTML]{F2CFC2}0.8874 & \cellcolor[HTML]{F2CFC2}89.71\% & \cellcolor[HTML]{F2CFC2}\textbf{4.84} & \cellcolor[HTML]{F2CFC2}0.8315\\

          & \cellcolor[HTML]{DDF6D2}\textsf{sw} & \cellcolor[HTML]{DDF6D2}\underline{38.31\%} & \cellcolor[HTML]{DDF6D2}24.04 & \cellcolor[HTML]{DDF6D2}\textbf{0.9141} & \cellcolor[HTML]{DDF6D2}84.80\% & \cellcolor[HTML]{DDF6D2}22.57 & \cellcolor[HTML]{DDF6D2}\textbf{0.8816}\\

            \midrule
            
          \multirow{6}{*}{\rotatebox[origin=c]{90}{\scriptsize{\lm{Gemma3-27B}}}}
          & \cellcolor[HTML]{d8d8d8}\textsf{en} & \cellcolor[HTML]{d8d8d8}\textbf{43.37\%} & \cellcolor[HTML]{d8d8d8}\underline{38.26} & \cellcolor[HTML]{d8d8d8}0.8542 & \cellcolor[HTML]{d8d8d8}\textbf{87.75\%} & \cellcolor[HTML]{d8d8d8}53.66 & \cellcolor[HTML]{d8d8d8}0.6275\\
          
          & \cellcolor[HTML]{E5CCFF}\textsf{ar} & \cellcolor[HTML]{E5CCFF}37.59\% & \cellcolor[HTML]{E5CCFF}36.32 & \cellcolor[HTML]{E5CCFF}\underline{0.8415}& \cellcolor[HTML]{E5CCFF}87.75\% & \cellcolor[HTML]{E5CCFF}\underline{81.96} & \cellcolor[HTML]{E5CCFF}\underline{0.4967} \\
          
          & \cellcolor[HTML]{E0E6FF}\textsf{de} & \cellcolor[HTML]{E0E6FF}38.19\% & \cellcolor[HTML]{E0E6FF}33.69 & \cellcolor[HTML]{E0E6FF}0.8633 & \cellcolor[HTML]{E0E6FF}\underline{79.41\%} & \cellcolor[HTML]{E0E6FF}34.94 & \cellcolor[HTML]{E0E6FF}0.6658\\

          & \cellcolor[HTML]{FFF1D5}\textsf{es} & \cellcolor[HTML]{FFF1D5}39.92\% & \cellcolor[HTML]{FFF1D5}31.18 & \cellcolor[HTML]{FFF1D5}0.8596 & \cellcolor[HTML]{FFF1D5}80.88\% & \cellcolor[HTML]{FFF1D5}30.73 & \cellcolor[HTML]{FFF1D5}0.6626\\

          & \cellcolor[HTML]{F2CFC2}\textsf{hi} & \cellcolor[HTML]{F2CFC2}36.43\% & \cellcolor[HTML]{F2CFC2}\textbf{11.30} & \cellcolor[HTML]{F2CFC2}0.8451& \cellcolor[HTML]{F2CFC2}81.37\% & \cellcolor[HTML]{F2CFC2}\textbf{4.35} & \cellcolor[HTML]{F2CFC2}0.6154\\

          & \cellcolor[HTML]{DDF6D2}\textsf{sw} & \cellcolor[HTML]{DDF6D2}\underline{33.90\%} & \cellcolor[HTML]{DDF6D2}23.30 & \cellcolor[HTML]{DDF6D2}\textbf{0.8731} & \cellcolor[HTML]{DDF6D2}87.25\% & \cellcolor[HTML]{DDF6D2}16.70 & \cellcolor[HTML]{DDF6D2}\textbf{0.7178}\\

            \midrule
          
          \multirow{6}{*}{\rotatebox[origin=c]{90}{\scriptsize{\lm{Llama3.3-70B}}}}          
          & \cellcolor[HTML]{d8d8d8}\textsf{en} & \cellcolor[HTML]{d8d8d8}\textbf{50.88\%} & \cellcolor[HTML]{d8d8d8}\underline{39.47} & \cellcolor[HTML]{d8d8d8}\underline{0.8429}	& \cellcolor[HTML]{d8d8d8}87.25\% & \cellcolor[HTML]{d8d8d8}52.84 & \cellcolor[HTML]{d8d8d8}0.6186\\
         
          & \cellcolor[HTML]{E5CCFF}\textsf{ar} & \cellcolor[HTML]{E5CCFF}36.91\% & \cellcolor[HTML]{E5CCFF}37.85 & \cellcolor[HTML]{E5CCFF}0.8626& \cellcolor[HTML]{E5CCFF}88.73\% & \cellcolor[HTML]{E5CCFF}\underline{77.32} & \cellcolor[HTML]{E5CCFF}\underline{0.4980}\\

          & \cellcolor[HTML]{E0E6FF}\textsf{de} & \cellcolor[HTML]{E0E6FF}42.25\% & \cellcolor[HTML]{E0E6FF}33.59 & \cellcolor[HTML]{E0E6FF}0.8689& \cellcolor[HTML]{E0E6FF}\underline{78.43\%} & \cellcolor[HTML]{E0E6FF}31.58 & \cellcolor[HTML]{E0E6FF}0.6385\\

          & \cellcolor[HTML]{FFF1D5}\textsf{es} &  \cellcolor[HTML]{FFF1D5}44.70\% &  \cellcolor[HTML]{FFF1D5}31.20 &  \cellcolor[HTML]{FFF1D5}0.8645	&  \cellcolor[HTML]{FFF1D5}83.33\% &  \cellcolor[HTML]{FFF1D5}29.41 &  \cellcolor[HTML]{FFF1D5}0.6567\\

          & \cellcolor[HTML]{F2CFC2}\textsf{hi} & \cellcolor[HTML]{F2CFC2}41.33\% & \cellcolor[HTML]{F2CFC2}\textbf{10.46} & \cellcolor[HTML]{F2CFC2}0.8476& \cellcolor[HTML]{F2CFC2}85.29\% & \cellcolor[HTML]{F2CFC2}\textbf{4.39} & \cellcolor[HTML]{F2CFC2}0.6182\\
          
          & \cellcolor[HTML]{DDF6D2}\textsf{sw} & \cellcolor[HTML]{DDF6D2}\underline{34.42\%} & \cellcolor[HTML]{DDF6D2}22.67 & \cellcolor[HTML]{DDF6D2}\textbf{0.8929}& \cellcolor[HTML]{DDF6D2}\textbf{91.18\%} & \cellcolor[HTML]{DDF6D2}14.43 & \cellcolor[HTML]{DDF6D2}\textbf{0.7792}\\

        \toprule[1.5pt]
        \end{tabular}
        }
        \caption{Directly generated counterfactuals $\tilde{x}_{\ell}$.}
        \label{subtab:direct}
  \end{subtable}
    
    &
    
    \begin{subtable}[c]{0.48\textwidth}
      \centering
      \resizebox{\textwidth}{!}{%
        \begin{tabular}{ccccc|ccc}

        \toprule[1.5pt]
         \multirow{2}{*}{\rotatebox[origin=c]{90}{\textbf{\scriptsize{Model}}}} & \textbf{Lang-} & \multicolumn{3}{c|}{\textbf{XNLI}} & \multicolumn{3}{c}{\textbf{SIB200}}\\ 
         
          & \textbf{uage} & LFR $\uparrow$ & PPL $\downarrow$ & TS $\uparrow$ & LFR $\uparrow$ & PPL $\downarrow$ & TS $\uparrow$\\
          
          \midrule

            \multirow{5}{*}{\rotatebox[origin=c]{90}{\scriptsize{\lm{Qwen2.5-7B}}}}
          & \cellcolor[HTML]{E5CCFF}\textsf{en}-\textsf{ar} & \cellcolor[HTML]{E5CCFF}43.49\% & \cellcolor[HTML]{E5CCFF}\underline{110.76} & \cellcolor[HTML]{E5CCFF}0.6897 & \cellcolor[HTML]{E5CCFF}90.20\% & \cellcolor[HTML]{E5CCFF}\underline{45.11} & \cellcolor[HTML]{E5CCFF}0.6669 \\

          & \cellcolor[HTML]{E0E6FF}\textsf{en}-\textsf{de} & \cellcolor[HTML]{E0E6FF}44.54\% & \cellcolor[HTML]{E0E6FF}73.59 & \cellcolor[HTML]{E0E6FF}\textbf{0.7838} & \cellcolor[HTML]{E0E6FF}\textbf{93.63\%} & \cellcolor[HTML]{E0E6FF}39.90 & \cellcolor[HTML]{E0E6FF}0.7491\\

          & \cellcolor[HTML]{FFF1D5}\textsf{en}-\textsf{es} & \cellcolor[HTML]{FFF1D5}\textbf{45.98\%} & \cellcolor[HTML]{FFF1D5}52.24 & \cellcolor[HTML]{FFF1D5}0.7826 & \cellcolor[HTML]{FFF1D5}92.16\% & \cellcolor[HTML]{FFF1D5}28.26 & \cellcolor[HTML]{FFF1D5}\textbf{0.7633} \\

          & \cellcolor[HTML]{F2CFC2}\textsf{en}-\textsf{hi} & \cellcolor[HTML]{F2CFC2}\underline{41.45\%} & \cellcolor[HTML]{F2CFC2}\textbf{9.40} & \cellcolor[HTML]{F2CFC2}0.6435 & \cellcolor[HTML]{F2CFC2}\underline{90.20\%} & \cellcolor[HTML]{F2CFC2}\textbf{9.19} & \cellcolor[HTML]{F2CFC2}0.6203 \\
        
          & \cellcolor[HTML]{DDF6D2}\textsf{en}-\textsf{sw} & \cellcolor[HTML]{DDF6D2}43.73\% & \cellcolor[HTML]{DDF6D2}57.39 & \cellcolor[HTML]{DDF6D2}\underline{0.2810} & \cellcolor[HTML]{DDF6D2}92.65\% & \cellcolor[HTML]{DDF6D2}46.35 & \cellcolor[HTML]{DDF6D2}\underline{0.2528}\\

            \midrule
            
          \multirow{5}{*}{\rotatebox[origin=c]{90}{\scriptsize{\lm{Gemma3-27B}}}}

          & \cellcolor[HTML]{E5CCFF}\textsf{en}-\textsf{ar} & \cellcolor[HTML]{E5CCFF}42.49\% & \cellcolor[HTML]{E5CCFF}48.27 & \cellcolor[HTML]{E5CCFF}0.6961 & \cellcolor[HTML]{E5CCFF}88.73\% & \cellcolor[HTML]{E5CCFF}\underline{27.09} & \cellcolor[HTML]{E5CCFF}0.5429 \\

          & \cellcolor[HTML]{E0E6FF}\textsf{en}-\textsf{de} & \cellcolor[HTML]{E0E6FF}42.49\% & \cellcolor[HTML]{E0E6FF}52.77 & \cellcolor[HTML]{E0E6FF}0.7629 & \cellcolor[HTML]{E0E6FF}\textbf{90.20\%} & \cellcolor[HTML]{E0E6FF}27.01 & \cellcolor[HTML]{E0E6FF}0.5753  \\

          & \cellcolor[HTML]{FFF1D5}\textsf{en}-\textsf{es} & \cellcolor[HTML]{FFF1D5}42.69\% & \cellcolor[HTML]{FFF1D5}\underline{50.43} & \cellcolor[HTML]{FFF1D5}\textbf{0.7692} & \cellcolor[HTML]{FFF1D5}89.22\% & \cellcolor[HTML]{FFF1D5}24.31 & \cellcolor[HTML]{FFF1D5}\textbf{0.5824}\\

          & \cellcolor[HTML]{F2CFC2}\textsf{en}-\textsf{hi} & \cellcolor[HTML]{F2CFC2}\underline{42.41\%} & \cellcolor[HTML]{F2CFC2}\textbf{5.73} & \cellcolor[HTML]{F2CFC2}0.7112 & \cellcolor[HTML]{F2CFC2}89.22\% & \cellcolor[HTML]{F2CFC2}\textbf{4.10} & \cellcolor[HTML]{F2CFC2}0.5451\\

          & \cellcolor[HTML]{DDF6D2}\textsf{en}-\textsf{sw} & \cellcolor[HTML]{DDF6D2}\textbf{43.01\%} & \cellcolor[HTML]{DDF6D2}28.28 &  \cellcolor[HTML]{DDF6D2}\underline{0.3569} & \cellcolor[HTML]{DDF6D2}\underline{85.78\%} & \cellcolor[HTML]{DDF6D2}13.66 & \cellcolor[HTML]{DDF6D2}\underline{0.2624}\\

            \midrule
          
          \multirow{5}{*}{\rotatebox[origin=c]{90}{\scriptsize{\lm{Llama3.3-70B}}}}          

          & \cellcolor[HTML]{E5CCFF}\textsf{en}-\textsf{ar} & \cellcolor[HTML]{E5CCFF}45.14\% & \cellcolor[HTML]{E5CCFF}\underline{169.00} & \cellcolor[HTML]{E5CCFF}0.6981 & \cellcolor[HTML]{E5CCFF}86.27\% & \cellcolor[HTML]{E5CCFF}\underline{34.47} & \cellcolor[HTML]{E5CCFF}\underline{0.5334} \\

          & \cellcolor[HTML]{E0E6FF}\textsf{en}-\textsf{de} & \cellcolor[HTML]{E0E6FF}47.58\% & \cellcolor[HTML]{E0E6FF}60.86 & \cellcolor[HTML]{E0E6FF}0.7627 & \cellcolor[HTML]{E0E6FF}\textbf{86.27\%} & \cellcolor[HTML]{E0E6FF}31.00 & \cellcolor[HTML]{E0E6FF}0.5854\\

          & \cellcolor[HTML]{FFF1D5}\textsf{en}-\textsf{es} & \cellcolor[HTML]{FFF1D5}\textbf{50.04\%} & \cellcolor[HTML]{FFF1D5}54.28 & \cellcolor[HTML]{FFF1D5}\textbf{0.7719} & \cellcolor[HTML]{FFF1D5}\underline{73.53\%} & \cellcolor[HTML]{FFF1D5}28.80 & \cellcolor[HTML]{FFF1D5}\textbf{0.5874} \\

          & \cellcolor[HTML]{F2CFC2}\textsf{en}-\textsf{hi} & \cellcolor[HTML]{F2CFC2}\underline{44.66\%} & \cellcolor[HTML]{F2CFC2}\textbf{5.51} & \cellcolor[HTML]{F2CFC2}0.7113 & \cellcolor[HTML]{F2CFC2}83.82\% & \cellcolor[HTML]{F2CFC2}\textbf{4.02} & \cellcolor[HTML]{F2CFC2}0.5441\\

          & \cellcolor[HTML]{DDF6D2}\textsf{en}-\textsf{sw} & \cellcolor[HTML]{DDF6D2}44.78\% & \cellcolor[HTML]{DDF6D2}38.65 & \cellcolor[HTML]{DDF6D2}\underline{0.3354} & \cellcolor[HTML]{DDF6D2}85.29\% & \cellcolor[HTML]{DDF6D2}13.53 & \cellcolor[HTML]{DDF6D2}\underline{0.2578}\\

        \toprule[1.5pt]
        \end{tabular}
        }
        \caption{Translation-based counterfactuals $\tilde{x}_{\textsf{en}\text{-}\ell}$.}
        \label{subtab:translation}
  \end{subtable}
    
    \end{tabular}
    
    \vspace{0.5em}
    \caption{Automatic evaluation results of counterfactuals based on label flip rate (LFR), perplexity (PPL), and textual similarity (TS) on \data{XNLI} and \data{SIB200} across \textit{English} (\colorbox[HTML]{d8d8d8}{\textsf{en}}), \textit{Arabic} (\colorbox[HTML]{E5CCFF}{\textsf{ar}}), \textit{German} (\colorbox[HTML]{E0E6FF}{\textsf{de}}), \textit{Spanish} (\colorbox[HTML]{FFF1D5}{\textsf{es}}), \textit{Hindi} (\colorbox[HTML]{F2CFC2}{\textsf{hi}}), and \textit{Swahili} (\colorbox[HTML]{DDF6D2}{\textsf{sw}}). \textbf{Bold}-faced languages indicate the best performance on a given metric, while \underline{underlined} languages denote the worst.
    }
    \label{tab:automatic_evaluation}
\end{table*}

%% file: tables/cda.tex
\begin{table}[ht!]
    \centering
    \renewcommand*{\arraystretch}{1}
    
    \footnotesize
    \resizebox{\columnwidth}{!}{%
 \begin{tabular}{ccccc|cc}

        \toprule[1.5pt]
         \multirow{2}{*}{\textbf{Model}} & \textbf{Counter} & \textbf{Lang} & \multicolumn{2}{c|}{\textbf{Cross-lingual}} & \multicolumn{2}{c}{\textbf{Multilingual}} \\ 



         & \textbf{-factual} & \textbf{-uage} & \textbf{XNLI} & \textbf{SIB200} & \textbf{XNLI} & \textbf{SIB200}\\
          \midrule
         \multirow{6}{*}{\rotatebox[origin=c]{90}{\large $\mathcal{M}_{base}$}} & - & \cellcolor[HTML]{d8d8d8}{\textsf{en}} & 68.70 & 83.80 & 72.22 & 82.83 \\
          & - & \cellcolor[HTML]{E5CCFF}{\textsf{ar}} & 60.12 & 25.30 & 63.21 & 54.55 \\
          & - & \cellcolor[HTML]{E0E6FF}{\textsf{de}} & 63.33 & 88.90 & 67.60 & 87.88 \\
          & - & \cellcolor[HTML]{FFF1D5}{\textsf{es}} & 66.05 & 87.90 & 68.72 & 87.88 \\
          & - & \cellcolor[HTML]{F2CFC2}{\textsf{hi}} & 56.09 & 74.70 & 62.04 & 80.81 \\
          & - & \cellcolor[HTML]{DDF6D2}{\textsf{sw}} & 48.66 & 64.60 & 59.00 & 78.79 \\

        \midrule

         \multirow{18}{*}{\rotatebox[origin=c]{90}{\large $\mathcal{M}_c$/$\mathcal{M}_m$}} & \centering \multirow{6}{*}{\rotatebox[origin=c]{90}{\lm{Qwen2.5-7B}}}  
          & \cellcolor[HTML]{d8d8d8}{\textsf{en}} & 69.86$_{\textcolor{deepgreen}{\text{+1.16}}}$ & 82.80$_{\textcolor{deepred}{\text{-1.00}}}$ & 73.45$_{\textcolor{deepgreen}{\text{+1.23}}}$ & 85.86$_{\textcolor{deepgreen}{\text{+3.03}}}$ \\
          && \cellcolor[HTML]{E5CCFF}{\textsf{ar}} & 58.10$_{\textcolor{deepred}{\text{-2.02}}}$ & 26.30$_{\textcolor{deepgreen}{\text{+1.00}}}$ & 64.89$_{\textcolor{deepgreen}{\text{+1.68}}}$ & 53.54$_{\textcolor{deepred}{\text{-1.01}}}$ \\
          && \cellcolor[HTML]{E0E6FF}{\textsf{de}} & 63.49$_{\textcolor{deepgreen}{\text{+0.16}}}$ & 84.80$_{\textcolor{deepred}{\text{-4.10}}}$ & 68.42$_{\textcolor{deepgreen}{\text{+0.82}}}$ & 84.85$_{\textcolor{deepred}{\text{-3.03}}}$ \\
          && \cellcolor[HTML]{FFF1D5}{\textsf{es}} & 65.43$_{\textcolor{deepred}{\text{-0.62}}}$ & 84.80$_{\textcolor{deepred}{\text{-3.10}}}$ & 69.94$_{\textcolor{deepgreen}{\text{+1.22}}}$ & 88.89$_{\textcolor{deepgreen}{\text{+1.01}}}$ \\
          && \cellcolor[HTML]{F2CFC2}{\textsf{hi}} & 55.33$_{\textcolor{deepred}{\text{-0.76}}}$ & 75.80$_{\textcolor{deepgreen}{\text{+1.10}}}$ & 62.32$_{\textcolor{deepgreen}{\text{+0.28}}}$ & 75.76$_{\textcolor{deepred}{\text{-5.05}}}$ \\
          && \cellcolor[HTML]{DDF6D2}{\textsf{sw}} & 48.92$_{\textcolor{deepgreen}{\text{+0.26}}}$ & 63.60$_{\textcolor{deepred}{\text{-1.00}}}$ & 57.74$_{\textcolor{deepred}{\text{-1.26}}}$ & 76.77$_{\textcolor{deepred}{\text{-2.02}}}$ \\

        \cmidrule(lr){2-7} 
            
         & \centering \multirow{6}{*}{\rotatebox[origin=c]{90}{\lm{Gemma3-27B}}}  
          & \cellcolor[HTML]{d8d8d8}{\textsf{en}} & 71.66$_{\textcolor{deepgreen}{\text{+2.96}}}$ & 85.90$_{\textcolor{deepgreen}{\text{+2.10}}}$ & 74.61$_{\textcolor{deepgreen}{\text{+2.39}}}$ & 86.87$_{\textcolor{deepgreen}{\text{+4.04}}}$ \\
          && \cellcolor[HTML]{E5CCFF}{\textsf{ar}} & 56.01$_{\textcolor{deepred}{\text{-4.11}}}$ & 23.20$_{\textcolor{deepred}{\text{-2.10}}}$ & 65.11$_{\textcolor{deepgreen}{\text{+1.90}}}$ & 49.49$_{\textcolor{deepred}{\text{-5.06}}}$ \\
          && \cellcolor[HTML]{E0E6FF}{\textsf{de}} & 62.53$_{\textcolor{deepred}{\text{-0.80}}}$ & 87.90$_{\textcolor{deepred}{\text{-1.00}}}$ & 68.66$_{\textcolor{deepgreen}{\text{+1.06}}}$ & 86.87$_{\textcolor{deepred}{\text{-1.01}}}$ \\
          && \cellcolor[HTML]{FFF1D5}{\textsf{es}} & 64.35$_{\textcolor{deepred}{\text{-1.70}}}$ & 86.90$_{\textcolor{deepred}{\text{-1.00}}}$ & 70.98$_{\textcolor{deepgreen}{\text{+2.26}}}$ & 89.90$_{\textcolor{deepgreen}{\text{+2.02}}}$ \\
          && \cellcolor[HTML]{F2CFC2}{\textsf{hi}} & 52.38$_{\textcolor{deepred}{\text{-3.71}}}$ & 73.70$_{\textcolor{deepred}{\text{-1.00}}}$ & 61.10$_{\textcolor{deepred}{\text{-0.94}}}$ & 83.84$_{\textcolor{deepgreen}{\text{+3.03}}}$ \\
          && \cellcolor[HTML]{DDF6D2}{\textsf{sw}} & 46.81$_{\textcolor{deepred}{\text{-1.85}}}$ & 64.60$_{\text{ 0.00}}$ & 55.57$_{\textcolor{deepred}{\text{-3.43}}}$ & 70.71$_{\textcolor{deepred}{\text{-8.08}}}$ \\

        \cmidrule(lr){2-7} 
          
           & \centering \multirow{6}{*}{\rotatebox[origin=c]{90}{\lm{Llama3.3-70B}}}           
          & \cellcolor[HTML]{d8d8d8}{\textsf{en}} & 70.86$_{\textcolor{deepgreen}{\text{+2.16}}}$ & 83.80$_{\text{ 0.00}}$ & 74.61$_{\textcolor{deepgreen}{\text{+2.39}}}$ & 83.84$_{\textcolor{deepgreen}{\text{+1.01}}}$ \\
          && \cellcolor[HTML]{E5CCFF}{\textsf{ar}} & 55.01$_{\textcolor{deepred}{\text{-5.11}}}$ & 25.30$_{\text{ 0.00}}$ & 64.77$_{\textcolor{deepgreen}{\text{+1.56}}}$ & 56.57$_{\textcolor{deepgreen}{\text{+2.02}}}$ \\
          && \cellcolor[HTML]{E0E6FF}{\textsf{de}} & 61.58$_{\textcolor{deepred}{\text{-1.75}}}$ & 83.80$_{\textcolor{deepred}{\text{-5.10}}}$ & 68.26$_{\textcolor{deepgreen}{\text{+0.66}}}$ & 87.88$_{\text{ 0.00}}$ \\
          && \cellcolor[HTML]{FFF1D5}{\textsf{es}} & 63.51$_{\textcolor{deepred}{\text{-2.54}}}$ & 84.80$_{\textcolor{deepred}{\text{-3.10}}}$ & 71.32$_{\textcolor{deepgreen}{\text{+2.60}}}$ & 88.89$_{\textcolor{deepgreen}{\text{+1.01}}}$ \\
          && \cellcolor[HTML]{F2CFC2}{\textsf{hi}} & 51.28$_{\textcolor{deepred}{\text{-4.81}}}$ & 73.70$_{\textcolor{deepred}{\text{-1.00}}}$ & 62.46$_{\textcolor{deepgreen}{\text{+0.42}}}$ & 79.80$_{\textcolor{deepred}{\text{-1.01}}}$ \\
          && \cellcolor[HTML]{DDF6D2}{\textsf{sw}} & 46.89$_{\textcolor{deepred}{\text{-1.77}}}$ & 59.60$_{\textcolor{deepred}{\text{-5.00}}}$ & 55.21$_{\textcolor{deepred}{\text{-3.79}}}$ & 73.74$_{\textcolor{deepred}{\text{-5.05}}}$ \\
        \toprule[1.5pt]
        \end{tabular}
    }
    \caption{Cross-lingual and multilingual CDA results (\textit{accuracy} in \%) for the base model $\mathcal{M}_{base}$ and the counterfactually augmented models $\mathcal{M}_c$ and $\mathcal{M}_m$.
    }
    \label{tab:cda}
\end{table}

%% file: tables/models.tex
\begin{table*}[t!]
    \centering
    \resizebox{\textwidth}{!}{%
        \begin{tabular}{cccc}

        \toprule
        \textbf{Name}& \textbf{Citation} & \textbf{Size} & \textbf{Link}\\

        \midrule
        \lm{Qwen2.5} & \citet{qwen2024qwen25technicalreport} & 7B & \url{https://huggingface.co/Qwen/Qwen2.5-7B-Instruct} \\
        \lm{Gemma3-27B} & \citet{gemmateam2025gemma3technicalreport} & 27B & \url{https://huggingface.co/google/gemma-3-27b-it}\\
        \lm{Llama3.3-70B} & \citet{grattafiori2024llama3herdmodels} & 70B & \url{https://huggingface.co/meta-llama/Llama-3.3-70B-Instruct} \\

        \bottomrule
        \end{tabular}
        }
    \caption{
    Three open-source LLMs are used for counterfactual generation.
    }
    \label{tab:used_model}
\end{table*}

%% file: tables/language.tex
\begin{table}[t!]
    \centering
    \resizebox{\columnwidth}{!}{%
        \begin{tabular}{cc}

        \toprule
        \textbf{Name}& \textbf{Language}\\

        \midrule
        \lm{Qwen2.5} & English, Spanish, German, Arabic\\
        \lm{Gemma3-27B} & n.a.\\
        \lm{Llama3.3-70B} & English, Hindi, Spanish, German \\

        
        
        
        
        \bottomrule
        \end{tabular}
        }
    \caption{
    Language support for the selected languages as shown in \S\ref{subsec:datasets}.
    }
    \label{tab:language}
\end{table}

%% file: tables/sib200_performance.tex
\begin{table}[t!]
    \centering
        \begin{tabular}{ccc}

        \toprule
        \textbf{Language} & \textbf{\data{SIB200}} & \textbf{\data{XNLI}}\\ 
        
        \midrule

        \textsf{en} & 87.75 & 81.57\\
        \textsf{de} & 86.27 & 71.53\\
        \textsf{ar} & 37.75 & 64.90\\
        \textsf{es} & 86.76 & 74.73\\
        \textsf{hi} & 78.43 & 59.88\\
        \textsf{sw} & 70.10 & 52.25\\
        
        \bottomrule
        \end{tabular}
        
    \caption{
    Task performance (in \%) of the explained \lm{mBERT} model across all selected languages on \data{SIB200} and \data{XNLI}.
    }
    \label{tab:sib200_performance}
\end{table}

%% file: tables/inference.tex
\begin{table}[h!]
    \centering
    \footnotesize
   
        \begin{tabular}{ccccc}

        \toprule
         \textbf{Model} & \textbf{\data{XNLI}} & \textbf{\data{SIB200}}\\
        \midrule

        \lm{Qwen2.5-7B} & 9h & 1h\\

        \lm{Gemma3-27B} & 11h & 8h\\

        \lm{Llama3.3-70B} & 17h & 13h\\

        \bottomrule
        \end{tabular}
    
    \caption{
    Inference time for counterfactual generation per language using \lm{Qwen2.5-7B}, \lm{Gemma3-27B}, and \lm{Llama3.3-70B} on \data{XNLI} and \data{SIB200}.
    }
    \label{tab:inference}
\end{table}

%% file: tables/automatic_translation.tex
\begin{table}[t!]
    \centering
    \renewcommand*{\arraystretch}{1}
    
    \footnotesize
    \resizebox{\columnwidth}{!}{%
 \begin{tabular}{ccccc|ccc}

        \toprule[1.5pt]
         \multirow{2}{*}{\textbf{Model}} & \textbf{Lang-} & \multicolumn{3}{c|}{\textbf{\data{XNLI}}} & \multicolumn{3}{c}{\textbf{\data{SIB200}}}\\

          & \textbf{uage} & \textbf{BLEU}   & \textbf{chrF}  & \textbf{\textls[-70]{XCOMET}}  & \textbf{BLEU}   & \textbf{chrF}  & \textbf{\textls[-70]{XCOMET}} \\
          
          \midrule

          \centering \multirow{5}{*}{\rotatebox[origin=c]{90}{\scriptsize{\lm{Qwen2.5-7B}}}} 
          & \textsf{en-ar} & 0.16 & 41.56 & 0.57 & 0.19 & \textbf{91.18} & 0.56\\
          & \textsf{en-de} & 0.25 & 54.37 & 0.69 & 0.30 & 90.20 & 0.70\\
          & \textsf{en-es} & \textbf{0.30} & \textbf{58.91} & \textbf{0.73} & \textbf{0.37} & \underline{87.75} & \textbf{0.76}\\
          & \textsf{en-hi} & 0.11 & 39.49 & 0.47 & 0.13 & 90.20 & 0.47\\
          & \textsf{en-sw} & \underline{0.04} & \underline{27.33} & \underline{0.43} & \underline{0.04} & 89.22 & \underline{0.42}\\

            \midrule
            
          \centering \multirow{5}{*}{\rotatebox[origin=c]{90}{\scriptsize{\lm{Gemma3-27B}}}}
          & \textsf{en-ar} & \underline{0.21} & 49.47 & 0.58 & \underline{0.19} & \underline{79.41} & 0.58\\
          & \textsf{en-de} & 0.25 & 52.13 & 0.71 & 0.22 & 79.90 & 0.72\\
          & \textsf{en-es} & \textbf{0.30} & \textbf{55.72} & \textbf{0.74} & \textbf{0.25} & 82.35 & \textbf{0.75} \\
          & \textsf{en-hi} & 0.23 & 50.17 & \underline{0.56} & 0.21 & 81.37 & \underline{0.55}\\
          & \textsf{en-sw} & 0.22 & \underline{48.41} & 0.60 & 0.19 & \textbf{83.82} & 0.59\\

            \midrule
          
          \centering \multirow{5}{*}{\rotatebox[origin=c]{90}{\scriptsize{\lm{Llama3.3-70B}}}}  
          & \textsf{en-ar} & 0.24 & \underline{44.86} & 0.62 & 0.18 & 86.27 & \underline{0.60}\\
          & \textsf{en-de} & 0.29 & 57.37 & 0.71 & 0.22 & 87.25 & 0.71\\
          & \textsf{en-es} & \textbf{0.35} & \textbf{60.68} & \textbf{0.75} & \textbf{0.25} & \underline{84.31} & \textbf{0.76}\\
          & \textsf{en-hi} & 0.22 & 45.88 & \underline{0.60} & 0.16 & 87.75 & 0.60\\
          & \textsf{en-sw} & \underline{0.21} & 45.88 & 0.64 & \underline{0.15} & \textbf{91.67} & 0.63\\

        \toprule[1.5pt]
        \end{tabular}
    }
    \caption{Machine translation evaluation of translation-based counterfactuals $\tilde{x}_{\textsf{en}\text{-}\ell}$ using BLUE, chrF, and XCOMET on \data{XNLI} and \data{SIB200}.
    }
    \label{tab:automatic_evaluation_translation}
\end{table}

%% file: tables/judge_avg.tex
\begin{table}[t!]
    \centering
    \renewcommand*{\arraystretch}{1}
    
    \footnotesize
    \resizebox{\columnwidth}{!}{%
 \begin{tabular}{cccc}

        \toprule[1.5pt]
         \scriptsize{\textbf{Dataset}} & \textbf{Language} & \textbf{\data{XNLI}} & \textbf{\data{SIB200}}\\
        
          \midrule

         \centering \multirow{5}{*}{\rotatebox[origin=c]{90}{\scriptsize{\lm{Qwen2.5-7B}}}} 
          & \textsf{en-ar}  & 60.00 &  \textbf{95.00} \\
          & \textsf{en-de}  &  84.50 & 88.25 \\
          & \textsf{en-es}  & \textbf{87.50} & 91.10 \\
          & \textsf{en-hi}  & 23.60 & 71.00\\
          & \textsf{en-sw}  & \underline{11.23} & \underline{7.88}  \\

            \midrule
            
           \centering \multirow{5}{*}{\rotatebox[origin=c]{90}{\scriptsize{\lm{Gemma3-27B}}}} 
          & \textsf{en-ar} & \textbf{88.00} & \textbf{98.25} \\
          & \textsf{en-de} &  80.50 & 92.50 \\
          & \textsf{en-es} & \underline{77.00}  & 90.55 \\
          & \textsf{en-hi} & 84.50  &  90.50 \\
          & \textsf{en-sw} & 83.50 & \underline{89.60}  \\

            \midrule
          
           \centering \multirow{5}{*}{\rotatebox[origin=c]{90}{\scriptsize{\lm{Llama3.3-70B}}}}           
          & \textsf{en-ar} & \underline{70.25} &  98.50 \\
          & \textsf{en-de} &  \textbf{90.00} & 97.88 \\
          & \textsf{en-es} & 88.50 & \textbf{99.40} \\
          & \textsf{en-hi} &  87.20 & \underline{84.05} \\
          & \textsf{en-sw} & 79.53 &  86.68 \\

        \toprule[1.5pt]
        \end{tabular}
    }
    \caption{Average Direct Assessment (DA) scores of back-translated counterfactuals $\tilde{x}_{\textsf{en}\text{-}\ell}$ on \data{XNLI} and \data{SIB200}. \textbf{Bold}-faced languages indicate the best translation performance, while \underline{underlined} languages denote the worst.
    }
    \label{tab:human_evaluation}
\end{table}

%% file: tables/correlation.tex
\begin{table}[t!]
    \centering
    \renewcommand*{\arraystretch}{1}
    
    \footnotesize
    \resizebox{\columnwidth}{!}{%
 \begin{tabular}{c|cccc}

        \toprule[1.5pt]
         \multirow{2}{*}{\textbf{Metric}} & \multicolumn{2}{c}{\textbf{\data{XNLI}}} & \multicolumn{2}{c}{\textbf{\data{SIB200}}} \\

        & $\rho$ & $p$-value & $\rho$ & $p$-value\\
          \midrule
          BLEU & 0.6018 & 0.0176 & 0.4865 & 0.0659\\
          chrF & 0.7746 & 0.0007 & -0.4776 & 0.0718\\
          XCOMET & 0.5157 & 0.0491 & 0.4598 & 0.0847 \\
          
        \toprule[1.5pt]
        \end{tabular}
    }
    \caption{Spearman's rank correlation ($\rho$) between automatic evaluation metric results and human evaluation results.
    }
    \label{tab:correlation}
\end{table}

%% file: tables/translation_analysis.tex
\begin{table}[t!]
    \centering
    \renewcommand*{\arraystretch}{1}
    
    \footnotesize
 \begin{tabular}{ccc}

        \toprule[1.5pt]
         \textbf{Language} & \textbf{IAA} & \textbf{$p$-value} \\

          \midrule
        \textsf{ar} & 0.7558 & $2.93e^{-12}$\\
        \textsf{de} & 0.5142 & $2.64e^{-05}$\\
        \textsf{es} & 0.5940 & $1.84e^{-06}$\\
        \textsf{hi} & 0.7440 & $9.61e^{-12}$\\
        \textsf{sw} & 0.9005 & $1.23e^{-22}$\\

        \toprule[1.5pt]
        \end{tabular}
    \caption{Inter-annotator agreement scores and $p$-values across all languages apart from English.
    }
    \label{tab:translation_analysis}
\end{table}

%% file: tables/perplexity.tex
\begin{table}[t!]
    \centering
    \resizebox{\columnwidth}{!}{%
        \begin{tabular}{cc|cc}

        \toprule
        \multicolumn{2}{c}{\textbf{\data{XNLI}}} &  \multicolumn{2}{c}{\textbf{\data{SIB200}}}\\
        \textbf{Language}& \textbf{Perplexity} & \textbf{Language}& \textbf{Perplexity}\\

        \midrule

        \textsf{en} & \underline{104.34} & \textsf{en} & 45.10\\
        \textsf{ar} & 78.32 & \textsf{ar} & \underline{51.53}\\
        \textsf{de} & 82.04 & \textsf{de} & \textbf{33.59}\\
        \textsf{es} & 88.00 & \textsf{es} & 35.43\\
        \textsf{hi} & \textbf{66.93} & \textsf{hi} & 42.20\\
        \textsf{sw} & 82.77 & \textsf{sw} & 38.36\\

        \bottomrule
        \end{tabular}
        }
    \caption{
    Perplexity of data points across the selected languages from the \data{XNLI} and \data{SIB200} datasets.
    }
    \label{tab:perplexity}
\end{table}

%% file: tables/model_training_params.tex
\begin{table*}[t!]
    \centering
    \renewcommand*{\arraystretch}{1}
    
    \footnotesize
        \begin{tabular}{ccccc|cccc}
        \toprule[1.5pt]
        \multirow{2}{*}{\textbf{Model}} & \multicolumn{4}{c|}{\textbf{Cross-lingual}} & \multicolumn{4}{c}{\textbf{Multilingual}} \\
        
        & \textbf{Size} & \textbf{Epochs} & \textbf{Batch} & \textbf{LR} & \textbf{Size} & \textbf{Epochs} & \textbf{Batch} & \textbf{LR} \\
        \midrule
        $\mathcal{M}_{base}$ & 2400 & 8 & 16 & $1e^{-05}$ & 2400 & 8 & 16 & $1e^{-05}$ \\
        \midrule
        $\mathcal{M}_{c}$/$\mathcal{M}_{m}$ (\lm{Gemma3-27B}) & 2400 & 8 & 16 & $1e^{-05}$ & 2400 & 8 & 16 & $1e^{-05}$ \\
        \midrule
        $\mathcal{M}_{c}$/$\mathcal{M}_{m}$ (\lm{Llama3.3-70B}) & 2400 & 12 & 24 & $2e^{-05}$ & 2000 & 12 & 8 & $2e^{-05}$ \\
        \midrule
        $\mathcal{M}_{c}$/$\mathcal{M}_{m}$ (\lm{Qwen2.5-7B}) & 2400 & 8 & 24 & $3e^{-05}$ & 2400 & 8 & 24 & $3e^{-05}$ \\
        \toprule[1.5pt]
        \end{tabular}
        
    \caption{Training configurations for \data{XNLI} models identified through grid search. Size = Training Size, Batch = Batch Size, LR = Learning Rate.}
    \label{tab:model_training_params}
\end{table*}

%% file: tables/cda_human.tex
\begin{table}[t!]
    \centering
    \renewcommand*{\arraystretch}{1}
    
    \footnotesize
    \resizebox{\columnwidth}{!}{%
 \begin{tabular}{ccccc|cc}

        \toprule[1.5pt]
         \multirow{2}{*}{\textbf{Model}} & \textbf{Counter} & \textbf{Lang} & \multicolumn{2}{c|}{\textbf{Cross-lingual}} & \multicolumn{2}{c}{\textbf{Multilingual}} \\

         & \textbf{-factual} & \textbf{-age} & \textbf{XNLI} & \textbf{SIB200} & \textbf{XNLI} & \textbf{SIB200}\\
          \midrule
         \multirow{6}{*}{$\mathcal{M}_{base}$} & - & \cellcolor[HTML]{d8d8d8}{\textsf{en}} & 38 & 68 & 38 & 78\\
          & - & \cellcolor[HTML]{E5CCFF}{\textsf{ar}} & 42 & 76 & 40 & 86\\
          & - & \cellcolor[HTML]{E0E6FF}{\textsf{de}}  & 44 & 72 & 40 & 78\\
          & - & \cellcolor[HTML]{FFF1D5}{\textsf{es}}  & 40 & 72 & 38 & 76\\
          & - & \cellcolor[HTML]{F2CFC2}{\textsf{hi}}  & 30 & 82 & 30 & 82\\
          & - & \cellcolor[HTML]{DDF6D2}{\textsf{sw}} & 42 & 48 & 38 & 62\\

        \midrule

         \multirow{18}{*}{$\mathcal{M}_c$/$\mathcal{M}_m$} & \centering \multirow{6}{*}{\rotatebox[origin=c]{90}{\lm{Qwen2.5-7B}}}  
          & \cellcolor[HTML]{d8d8d8}{\textsf{en}} &   26$_{\textcolor{deepred}{\text{-12}}}$ & 82$_{\textcolor{deepgreen}{\text{+14}}}$ & 36$_{\textcolor{deepred}{\text{-2}}}$ & 80$_{\textcolor{deepgreen}{\text{+2}}}$ \\
          &  & \cellcolor[HTML]{E5CCFF}{\textsf{ar}} & 36$_{\textcolor{deepred}{\text{-7}}}$ & 78$_{\textcolor{deepgreen}{\text{+2}}}$ & 48$_{\textcolor{deepgreen}{\text{+8}}}$ & 86$_{\textcolor{black}{\text{0}}}$ \\
          &  & \cellcolor[HTML]{E0E6FF}{\textsf{de}}  & 34$_{\textcolor{deepred}{\text{-10}}}$ & 74$_{\textcolor{deepgreen}{\text{+2}}}$ & 40$_{\textcolor{black}{\text{0}}}$ & 82$_{\textcolor{deepgreen}{\text{+4}}}$ \\
          &  & \cellcolor[HTML]{FFF1D5}{\textsf{es}}  & 36$_{\textcolor{deepred}{\text{-4}}}$ & 82$_{\textcolor{deepgreen}{\text{+10}}}$ & 42$_{\textcolor{deepgreen}{\text{+4}}}$ & 80$_{\textcolor{deepgreen}{\text{+4}}}$\\
          &  & \cellcolor[HTML]{F2CFC2}{\textsf{hi}}  & 26$_{\textcolor{deepred}{\text{-4}}}$ & 80$_{\textcolor{deepred}{\text{-2}}}$ & 30$_{\textcolor{black}{\text{0}}}$ & 86$_{\textcolor{deepgreen}{\text{+4}}}$\\
          &  & \cellcolor[HTML]{DDF6D2}{\textsf{sw}} & 38$_{\textcolor{deepred}{\text{-4}}}$ & 52$_{\textcolor{deepgreen}{\text{+4}}}$ & 48$_{\textcolor{deepgreen}{\text{+10}}}$ & 60$_{\textcolor{deepred}{\text{-2}}}$\\

        \cmidrule(lr){2-7} 
            
         & \centering \multirow{6}{*}{\rotatebox[origin=c]{90}{\lm{Gemma3-27B}}}  
          & \cellcolor[HTML]{d8d8d8}{\textsf{en}} &   34$_{\textcolor{deepred}{\text{-4}}}$ & 86$_{\textcolor{deepgreen}{\text{+18}}}$ & 38$_{\textcolor{black}{\text{0}}}$ & 84$_{\textcolor{deepgreen}{\text{+6}}}$\\
          &  & \cellcolor[HTML]{E5CCFF}{\textsf{ar}} & 40$_{\textcolor{deepred}{\text{-2}}}$ & 78$_{\textcolor{deepgreen}{\text{+2}}}$ & 44$_{\textcolor{deepgreen}{\text{+4}}}$ & 88$_{\textcolor{deepgreen}{\text{+2}}}$ \\
          &  & \cellcolor[HTML]{E0E6FF}{\textsf{de}}  & 36$_{\textcolor{deepred}{\text{-8}}}$ & 80$_{\textcolor{deepgreen}{\text{+8}}}$ & 38$_{\textcolor{deepred}{\text{-2}}}$ & 84$_{\textcolor{deepgreen}{\text{+6}}}$\\
          &  & \cellcolor[HTML]{FFF1D5}{\textsf{es}}  & 38$_{\textcolor{deepred}{\text{-2}}}$ & 84$_{\textcolor{deepgreen}{\text{+12}}}$ & 36$_{\textcolor{deepred}{\text{-2}}}$ & 82$_{\textcolor{deepgreen}{\text{+6}}}$\\
          &  & \cellcolor[HTML]{F2CFC2}{\textsf{hi}} &  32$_{\textcolor{deepgreen}{\text{+2}}}$ & 80$_{\textcolor{deepred}{\text{-2}}}$ & 24$_{\textcolor{deepred}{\text{-6}}}$ & 82$_{\textcolor{black}{\text{0}}}$\\
          &  & \cellcolor[HTML]{DDF6D2}{\textsf{sw}} & 36$_{\textcolor{deepred}{\text{-6}}}$ & 52$_{\textcolor{deepgreen}{\text{+4}}}$ & 38$_{\textcolor{black}{\text{0}}}$ & 60$_{\textcolor{deepred}{\text{-2}}}$\\

        \cmidrule(lr){2-7} 
          
           & \centering \multirow{6}{*}{\rotatebox[origin=c]{90}{\lm{Llama3.3-70B}}}                     
          & \cellcolor[HTML]{d8d8d8}{\textsf{en}} &   34$_{\textcolor{deepred}{\text{-4}}}$ & 82$_{\textcolor{deepgreen}{\text{+14}}}$ & 30$_{\textcolor{deepred}{\text{-8}}}$ & 82$_{\textcolor{deepgreen}{\text{+4}}}$\\
          &  & \cellcolor[HTML]{E5CCFF}{\textsf{ar}} & 42$_{\textcolor{black}{\text{0}}}$ & 80$_{\textcolor{deepgreen}{\text{+4}}}$ & 46$_{\textcolor{deepgreen}{\text{+6}}}$ & 86$_{\textcolor{black}{\text{0}}}$\\
          &  & \cellcolor[HTML]{E0E6FF}{\textsf{de}}  & 46$_{\textcolor{deepgreen}{\text{+2}}}$ & 80$_{\textcolor{deepgreen}{\text{+8}}}$ & 32$_{\textcolor{deepred}{\text{-8}}}$ & 80$_{\textcolor{deepgreen}{\text{+2}}}$\\
          &  & \cellcolor[HTML]{FFF1D5}{\textsf{es}}  & 40$_{\textcolor{black}{\text{0}}}$ & 78$_{\textcolor{deepgreen}{\text{+6}}}$ & 34$_{\textcolor{deepred}{\text{-4}}}$ & 78$_{\textcolor{deepgreen}{\text{+2}}}$\\
          &  & \cellcolor[HTML]{F2CFC2}{\textsf{hi}}  & 36$_{\textcolor{deepgreen}{\text{+6}}}$ & 80$_{\textcolor{deepred}{\text{-2}}}$ & 38$_{\textcolor{deepgreen}{\text{+8}}}$ & 88$_{\textcolor{deepgreen}{\text{+6}}}$\\
          &  & \cellcolor[HTML]{DDF6D2}{\textsf{sw}} & 44$_{\textcolor{deepgreen}{\text{+2}}}$ & 46$_{\textcolor{deepred}{\text{-2}}}$ & 42$_{\textcolor{deepgreen}{\text{+4}}}$ & 52$_{\textcolor{deepred}{\text{-10}}}$\\
        \toprule[1.5pt]
        \end{tabular}
    }
    \caption{Cross-lingual and multilingual CDA results (in \%) for the base model $\mathcal{M}_{base}$ and the counterfactually augmented models $\mathcal{M}_c$ and $\mathcal{M}_m$ using directly generated counterfactuals $\tilde{x}_\ell$ on \data{XNLI} and \data{SIB200}.
    }
    \label{tab:cda_human}
\end{table}

%% file: tables/cda_translation.tex
\begin{table}[t!]
    \centering
    \renewcommand*{\arraystretch}{1}
    
    \footnotesize
    \resizebox{\columnwidth}{!}{%
 \begin{tabular}{ccccccc}

        \toprule[1.5pt]

         
         \textbf{Model} & \textbf{Counter} & \textbf{Lang} & \multicolumn{2}{c}{\textbf{Test set}} & \multicolumn{2}{c}{\textbf{Human}} \\

         \textbf{Dataset} & \textbf{-factual} & \textbf{-age} & \textbf{XNLI} & \textbf{SIB200} & \textbf{XNLI} & \textbf{SIB200}\\
          \midrule

         \multirow{6}{*}{$\mathcal{M}_{base}$} 
        & - & \cellcolor[HTML]{d8d8d8}{\textsf{en}} &  72.22 & 82.83 & 38 & 78   \\
          & - & \cellcolor[HTML]{E5CCFF}{\textsf{ar}}  & 63.21 & 54.55 & 40 & 86\\
          & - & \cellcolor[HTML]{E0E6FF}{\textsf{de}} & 67.60 & 87.88 & 40 & 78\\
          & - & \cellcolor[HTML]{FFF1D5}{\textsf{es}}  & 68.72 & 87.88 & 38 & 76\\
          & - & \cellcolor[HTML]{F2CFC2}{\textsf{hi}}   & 62.04 & 80.81 & 30 & 82\\
          & - & \cellcolor[HTML]{DDF6D2}{\textsf{sw}} & 59.00 & 78.79 & 38 & 62\\

        \midrule

         \multirow{18}{*}{$\mathcal{M}_m$} & \centering \multirow{6}{*}{\rotatebox[origin=c]{90}{\lm{Qwen2.5-7B}}}  
          & \cellcolor[HTML]{d8d8d8}{\textsf{en}} & 70.66$_{\textcolor{deepred}{\text{-1.56}}}$ & 83.84$_{\textcolor{deepgreen}{\text{+1.01}}}$ & 40$_{\textcolor{deepgreen}{\text{+2}}}$ & 76$_{\textcolor{deepred}{\text{-2}}}$  \\
          &  & \cellcolor[HTML]{E5CCFF}{\textsf{ar}} & 63.41$_{\textcolor{deepgreen}{\text{+0.2}}}$ & 54.55$_{\textcolor{black}{\text{0.00}}}$ & 32$_{\textcolor{deepred}{\text{-8}}}$ & 78$_{\textcolor{deepred}{\text{-8}}}$  \\
          &  & \cellcolor[HTML]{E0E6FF}{\textsf{de}} & 67.11$_{\textcolor{deepred}{\text{-0.49}}}$ & 83.84$_{\textcolor{deepred}{\text{-4.04}}}$ & 42$_{\textcolor{deepgreen}{\text{+2}}}$ & 78$_{\textcolor{black}{\text{0}}}$ \\
          &  & \cellcolor[HTML]{FFF1D5}{\textsf{es}} & 67.96$_{\textcolor{deepred}{\text{-0.76}}}$ & 87.88$_{\textcolor{black}{\text{0.00}}}$ & 38$_{\textcolor{black}{\text{0}}}$ & 76$_{\textcolor{black}{\text{0}}}$ \\
          &  & \cellcolor[HTML]{F2CFC2}{\textsf{hi}} & 61.58$_{\textcolor{deepred}{\text{-0.46}}}$ & 81.82$_{\textcolor{deepgreen}{\text{+1.01}}}$ & 34$_{\textcolor{deepgreen}{\text{+4}}}$ & 78$_{\textcolor{deepred}{\text{-4}}}$ \\
          &  & \cellcolor[HTML]{DDF6D2}{\textsf{sw}} & 57.80$_{\textcolor{deepred}{\text{-1.2}}}$ & 70.71$_{\textcolor{deepred}{\text{-8.08}}}$ & 36$_{\textcolor{deepred}{\text{-2}}}$ & 52$_{\textcolor{deepred}{\text{-10}}}$ \\

            \cmidrule(lr){2-7} 
            
         & \centering \multirow{6}{*}{\rotatebox[origin=c]{90}{\lm{Gemma3-27B}}}  & \cellcolor[HTML]{d8d8d8}{\textsf{en}} &
         70.18$_{\textcolor{deepred}{\text{-2.04}}}$ & 88.89$_{\textcolor{deepgreen}{\text{+6.06}}}$ & 40$_{\textcolor{deepgreen}{\text{+2}}}$  & 82$_{\textcolor{deepgreen}{\text{+4}}}$ \\
          &  & \cellcolor[HTML]{E5CCFF}{\textsf{ar}} & 63.75$_{\textcolor{deepgreen}{\text{+0.54}}}$ & 51.52$_{\textcolor{deepred}{\text{-3.03}}}$ & 34$_{\textcolor{deepred}{\text{-6}}}$ & 88$_{\textcolor{deepgreen}{\text{+2}}}$  \\
          &  & \cellcolor[HTML]{E0E6FF}{\textsf{de}} & 66.63$_{\textcolor{deepred}{\text{-0.97}}}$ & 85.86$_{\textcolor{deepred}{\text{-2.02}}}$ & 46$_{\textcolor{deepgreen}{\text{+6}}}$ & 84$_{\textcolor{deepgreen}{\text{+6}}}$ \\
          &  & \cellcolor[HTML]{FFF1D5}{\textsf{es}} &  67.45$_{\textcolor{deepred}{\text{-1.27}}}$ & 88.89$_{\textcolor{deepgreen}{\text{+1.01}}}$ & 38$_{\textcolor{black}{\text{0}}}$ & 82$_{\textcolor{deepgreen}{\text{+6}}}$ \\
          &  & \cellcolor[HTML]{F2CFC2}{\textsf{hi}} & 61.18$_{\textcolor{deepred}{\text{-0.86}}}$ & 79.80$_{\textcolor{deepred}{\text{-1.01}}}$  & 34$_{\textcolor{deepgreen}{\text{+4}}}$ & 88$_{\textcolor{deepgreen}{\text{+6}}}$ \\
          &  & \cellcolor[HTML]{DDF6D2}{\textsf{sw}} & 58.64$_{\textcolor{deepred}{\text{-0.36}}}$ & 77.78$_{\textcolor{deepred}{\text{-1.01}}}$  & 42$_{\textcolor{deepgreen}{\text{+4}}}$ & 70$_{\textcolor{deepgreen}{\text{+8}}}$ \\

            \cmidrule(lr){2-7} 
          
          & \centering \multirow{6}{*}{\rotatebox[origin=c]{90}{\lm{Llama3.3-70B}}}           
         & \cellcolor[HTML]{d8d8d8}{\textsf{en}} & 71.26$_{\textcolor{deepred}{\text{-0.96}}}$ & 87.88$_{\textcolor{deepgreen}{\text{+5.05}}}$ & 40$_{\textcolor{deepgreen}{\text{+2}}}$ & 78$_{\textcolor{black}{\text{0}}}$  \\
          &  & \cellcolor[HTML]{E5CCFF}{\textsf{ar}} & 64.45$_{\textcolor{deepgreen}{\text{+1.24}}}$ & 52.53$_{\textcolor{deepred}{\text{-2.02}}}$ & 38$_{\textcolor{deepred}{\text{-2}}}$ & 88$_{\textcolor{deepgreen}{\text{+2}}}$  \\
          &  & \cellcolor[HTML]{E0E6FF}{\textsf{de}} & 67.47$_{\textcolor{deepred}{\text{-0.13}}}$ & 87.88$_{\textcolor{black}{\text{0.00}}}$ & 38$_{\textcolor{deepred}{\text{-2}}}$ & 84$_{\textcolor{deepgreen}{\text{+6}}}$ \\
          &  & \cellcolor[HTML]{FFF1D5}{\textsf{es}} &  69.36$_{\textcolor{deepgreen}{\text{+0.64}}}$ & 86.87$_{\textcolor{deepred}{\text{-1.01}}}$ & 38$_{\textcolor{black}{\text{0}}}$ & 76$_{\textcolor{black}{\text{0}}}$ \\
          &  & \cellcolor[HTML]{F2CFC2}{\textsf{hi}} & 61.25$_{\textcolor{deepred}{\text{-0.79}}}$ & 76.77$_{\textcolor{deepred}{\text{-4.04}}}$ & 28$_{\textcolor{deepred}{\text{-2}}}$ & 82$_{\textcolor{black}{\text{0}}}$ \\
          &  & \cellcolor[HTML]{DDF6D2}{\textsf{sw}} & 58.12$_{\textcolor{deepred}{\text{-0.88}}}$ & 72.73$_{\textcolor{deepred}{\text{-6.06}}}$ & 40$_{\textcolor{deepgreen}{\text{+2}}}$ & 74$_{\textcolor{deepgreen}{\text{+12}}}$ \\

        \toprule[1.5pt]
        \end{tabular}
    }
    \caption{CDA results (in \%) for the base model $\mathcal{M}_{base}$ and the counterfactually augmented model $\mathcal{M}_m$ using translation-based counterfactuals $\tilde{x}_{\textsf{en}-\ell}$ on \data{XNLI} and \data{SIB200}.
    }
    \label{tab:cda_translation}
\end{table}

%% file: tables/cda_error_analysis.tex
\begin{table}[t!]
    \centering
    \begin{tabular}{cc}
        \toprule
        \textbf{Language} & \textbf{CDA Performance} \\
        \midrule
        \textsf{en}-before & 73.45 \\
        \textsf{en}-after & 73.62 (+0.17) \\
        \midrule
        \textsf{ar}-before & 64.89 \\
        \textsf{ar}-after & 65.26 (+0.37) \\
        \midrule
        \textsf{de}-before & 68.42 \\
        \textsf{de}-after & 69.07 (+0.65) \\
        \midrule
        \textsf{es}-before & 69.94 \\
        \textsf{es}-after & 71.12 (+1.18) \\
        \midrule
        \textsf{hi}-before & 75.76 \\
        \textsf{hi}-after & 78.10 (+2.34) \\
        \midrule
        \textsf{sw}-before & 76.77 \\
        \textsf{sw}-after & 78.92 (+2.15) \\
        \bottomrule
    \end{tabular}
    \caption{Counterfactual data augmentation (CDA) performance comparison before and after filtering out error cases (\textit{copy-paste} and \textit{language confusion}).} 
    \label{tab:cda_error_analysis}
\end{table}